%% file: main.tex
\documentclass[10pt]{article} 
\usepackage[accepted,preprint]{tmlr}

\input{math_commands.tex}

\usepackage{hyperref}
\usepackage{url}
\usepackage{amssymb}

\usepackage{graphicx}
\usepackage{algorithm}
\usepackage{algpseudocode}
\usepackage{comment}
\usepackage{nicefrac}
\usepackage{threeparttable} 
\usepackage{tabularx}
\usepackage{xfp}
\usepackage{pifont}

\usepackage[normalem]{ulem} 

\usepackage{booktabs}

\usepackage[normalem]{ulem} 
\definecolor{green}{RGB}{0, 180, 0}
\definecolor{red}{RGB}{220, 0, 0}

\newcommand{\cmark}{\textcolor{green}{\checkmark}}
\newcommand{\xmark}{\textcolor{red}{\ding{55}}}

\providecommand{\rebuttal}[1]{#1}

\title{The Last Mile to Supervised Performance: \\Semi-Supervised Domain Adaptation for Semantic Segmentation}

\author{\name Daniel Morales-Brotons \email danimoralesbrotons@gmail.com \\
      \addr EPFL
      \AND
      \name Grigorios Chrysos \thanks{Work done while at EPFL} \email chrysos@wisc.edu \\
      \addr University of Wisconsin-Madison
      \AND
      \name Stratis Tzoumas \email stratis.tzoumas@zeiss.com \\
      \addr ZEISS
      \AND
      \name Volkan Cevher \email volkan.cevher@epfl.ch \\
      \addr EPFL
}

\def\month{03}  
\def\year{2024} 
\def\openreview{\url{https://openreview.net/forum?id=2M9CUnYnBA}} 

\begin{document}

\maketitle

\begin{abstract}
Supervised deep learning requires massive labeled datasets, but obtaining annotations is not always easy or possible, especially for dense tasks like semantic segmentation. To overcome this issue, numerous works explore Unsupervised Domain Adaptation (UDA), which uses a labeled dataset from another domain (source), or Semi-Supervised Learning (SSL), which trains on a partially labeled set. Despite the success of UDA and SSL, reaching supervised performance at a low annotation cost remains a notoriously elusive goal. To address this, we study the promising setting of Semi-Supervised Domain Adaptation (SSDA). We propose a simple SSDA framework that combines consistency regularization, pixel contrastive learning, and self-training to effectively utilize a few target-domain labels. Our method outperforms prior art in the popular GTA$\rightarrow$Cityscapes benchmark and shows that as little as $50$ target labels can suffice to achieve near-supervised performance. Additional results on Synthia$\rightarrow$Cityscapes, GTA$\rightarrow$BDD and Synthia$\rightarrow$BDD further demonstrate the effectiveness and practical utility of the method. Lastly, we find that existing UDA and SSL methods are not well-suited for the SSDA setting and discuss design patterns to adapt them.

\end{abstract}

\section{Introduction}

Semantic segmentation is a key task in computer vision with diverse applications ranging from autonomous driving \citep{badrinarayanan2017segnet} to medical image analysis \citep{ronneberger2015u}. 
Despite recent progress in this area using supervised learning methods \citep{badrinarayanan2017segnet, ronneberger2015u, xie2021segformer}, supervision remains challenging in practical applications due to the high labeling cost and the need for specialized domain experts. Therefore, minimizing the labeling cost while maintaining strong performance is critical. 
Common approaches for learning with unlabeled data are Unsupervised Domain Adaptation (UDA), which uses additional data from another similar domain, and Semi-Supervised Learning (SSL), which trains on a partially labeled set. 

\begin{figure}[!b]
    \centering
    \includegraphics[width=0.57\linewidth]{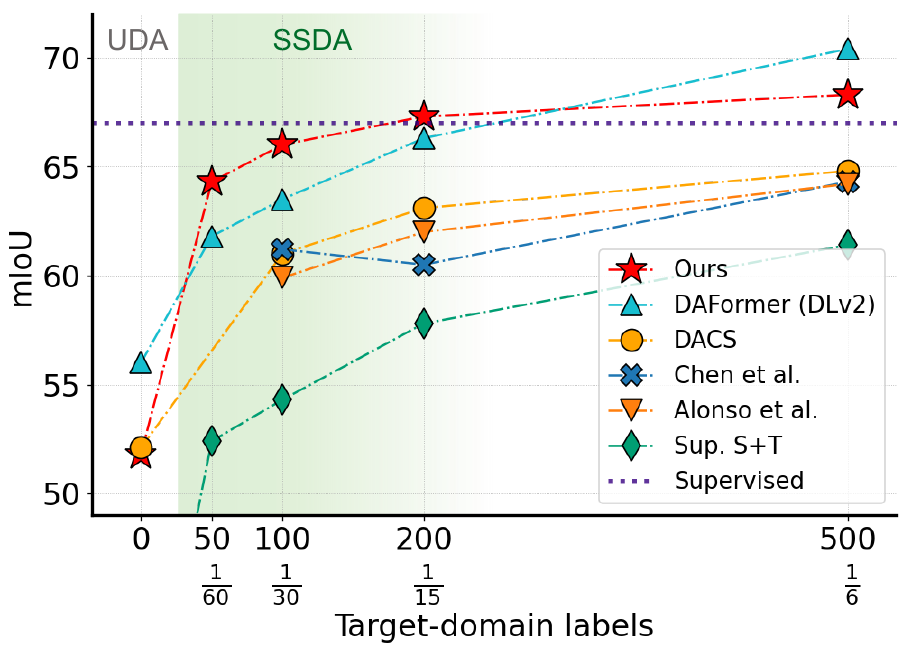}
    \caption{GTA$\rightarrow$Cityscapes results (mIoU). Our method beats all baselines in the highlighted regime of interest: SSDA with a low amount of target labels. We claim SSDA as an alternative to UDA where near-supervised performance can be achieved at a low annotation cost. ``Supervised" indicates a model trained on the full target dataset (2975 images). Fractions represent ratio of target-domain samples labeled. Results are an average of 3 runs on a DeepLabv2 + ResNet-101 network. See Tab. \ref{tab:SSDA_sota} for the results table.
    }
    \label{fig:UDA2SSDA}
\end{figure}

While UDA has demonstrated promising results on public benchmarks, its practical implementation remains challenging. Although UDA methods do not require target annotations for training and leverage additional labeled data from a source domain, they often require target labels for hyperparameter tuning~\citep{saito2021tune}. Moreover, it is essential in industrial and medical applications to have a well-validated system, which necessitates the collection of a target labeled set for validation purposes. In such cases, annotating a few samples for training may not a significant overhead. Another setting to learn with missing labels is SSL, which trains a model on a partially labeled dataset~\citep{chen2021CPS, AlonsoSSL, olsson2021classmix}. However, SSL methods may underperform and risk overfitting when the number of labels is low. While adding a source dataset can alleviate this problem, existing SSL methods are not designed to leverage data from another domain, and studies like the one of \citet{AlonsoSSL} have shown only moderate improvement. Despite the competitive performance of both UDA and SSL methods, they fall short of supervised performance, as they achieve significantly lower accuracies than the fully supervised counterpart.

In this work, we study how to close the gap to supervised performance by exploiting the Semi-Supervised Domain Adaptation (SSDA) setting, and show that it is possible to match supervised accuracy at a modest annotation cost. SSDA is essentially the combination of SSL and UDA, as it uses source labeled data, target unlabeled data, and a few target labels (Tab. \ref{tab:settings}). Despite its practical value and performance potential while alleviating annotation requirements, SSDA has received less attention \citep{Berthelot_adamatch}. To our knowledge, only two works present a semantic segmentation method tailored to SSDA \citep{Wang_ASS, ChenDual}, and \citet{AlonsoSSL} propose an SSL method and try to extend it to SSDA. Moreover, the existing UDA works do not explore incorporating a few target labels and are suboptimal in an SSDA setting.

\begin{table}[!t]
    \begin{center}
    \caption{\textbf{Summary of settings.} Types of data used in Semi-Supervised Learning (SSL), Unsupervised Domain Adaptation (UDA) and Semi-Supervised Domain Adaptation (SSDA).}
    \vspace{3mm}
    \begin{tabular}{lccc}
        \toprule
        Data & \begin{tabular}{@{}c@{}} Source \\ Labeled \end{tabular} & \begin{tabular}{@{}c@{}} Target \\ Labeled \end{tabular} & \begin{tabular}{@{}c@{}} Target \\ Unlabeled \end{tabular} \\
        \midrule
        SSL & \xmark & \pmb\cmark & \pmb\cmark \\
        UDA & \pmb\cmark & \xmark & \pmb\cmark \\
        SSDA & \pmb{\cmark} & \pmb\cmark & \pmb\cmark\\
        \bottomrule
    \end{tabular}
    \end{center}
    \label{tab:settings}
\end{table}

We introduce a simple and straightforward semantic segmentation framework tailored to SSDA, which uses a combination of consistency regularization (CR) and pixel contrastive learning (PCL). The main goal of the method is to achieve compact clusters of target representations, which facilitate the classification task, while also learning a domain-robust feature extractor to leverage the source domain data. Moreover, we also focus on effectively utilizing the few available target labels. Finally, we propose a self-training scheme that improves training efficiency by iteratively refining model and pseudolabels. Our comprehensive evaluation on GTA$\rightarrow$Cityscapes demonstrates how the proposed method achieves state-of-the-art performance on SSDA semantic segmentation, approaching the supervised performance with minimal annotation, using $\le\nicefrac{1}{15}$ of target labels (see Fig. \ref{fig:UDA2SSDA}). Additional results in other benchmarks, without further hyperparameter tuning, confirm the effectiveness and high practical value of the method. We will make the code available upon acceptance.

The main contributions of this paper are:
    \begin{itemize}
        \item 
        We present a simple SSDA method for semantic segmentation that effectively utilizes the different kinds of data available, reaching a performance comparable to supervised learning.
    
        \item 
        We demonstrate a significant improvement of SSDA over UDA even with only 50 target labels ($+6.9$ mIoU). We also find that existing UDA methods are suboptimal in SSDA and discuss potential avenues for adapting them.
    
        \item 
        We investigate the relationship between SSL and SSDA, and show an improvement over the former ($+9.0$ mIoU at $50$ labels) when effectively leveraging source domain data.
    
    
    \end{itemize}

\section{Related Work}

\subsection{Unsupervised Domain Adaptation (UDA) for semantic segmentation} 
Numerous approaches have been proposed for UDA in semantic segmentation. 
In recent years, these techniques have been broadly classified into two main categories: adversarial training and self-training. Adversarial training methods minimize the difference between the source and target domains through a minimax game between a feature extractor and a domain discriminator \citep{Ganin_DANN, Hoffman_Cycada, Vu_AdvEnt, wang2020classes}. Conversely, self-training methods involve producing pseudolabels for the target domain data and aligning the two domains by means of domain mixing \citep{Tranheden_dacs} or source styling \citep{Yang_FDA}. Pseudolabels can be carefully generated using prototypes \citep{Zhang_proda, liu2021bapa} or adaptive confidence thresholds \citep{Mei_IAST}. An iterative self-training algorithm that employs pseudolabels is explored by \citet{zou2018cbst} and \citet{Li2019bidirectional}. While all the above-mentioned methods employ the Deeplab family of architectures \citep{deeplabv2, deeplabv3}, recent studies have shown that self-training methods using Transformer-based networks have achieved state-of-the-art performance \citep{Hoyer_DAFormer, Hoyer_HRDA}. Lastly, \citet{Hoyer_HRDA} utilize high resolution and multi-scale inputs with a module that can be applied on top of existing UDA methods. \rebuttal{Even though several ideas from UDA can potentially be employed in SSDA frameworks, out-of-the-box UDA methods are suboptimal in SSDA (see \ref{sec:uda2ssda}), since they do not consider how to fully leverage the few, very valuable, target labels. Therefore, to fully leverage the provided labels, we need to design frameworks tailored to SSDA.} 

\subsection{Semi-Supervised Learning (SSL) for semantic segmentation} 
Learning on a partially labeled dataset has been largely explored for semantic segmentation. A commonly used mechanism is consistency regularization, which aims to learn a model invariant to perturbations by encouraging consistent predictions between augmentations of an unlabeled image. Relying on the cluster and smoothness assumptions \citep{cluster_assumption}, it encourages compact clusters of representations separated by low-density regions, where the decision boundary can lie. Some approaches use a mean teacher to generate pseudolabels \citep{MeanTeacher, french2019cutmix, AlonsoSSL, liu2022perturbed}, while others train a single model \citep{sohn2020fixmatch, zou2020pseudoseg} or perform cross-supervision between two models \citep{chen2021CPS, fan2022ucc, ke2020guided}. 

Iterative self-training consists of training for one round and using the resulting model to generate pseudolabels to train a new model in the next round \citep{xie2020noisy_student, zoph2020rethinking, zou2020pseudoseg, teh2022gist, liu2022adaptive}. In contrast to consistency regularization, the pseudolabels are generated offline. Despite its effectiveness, self-training can suffer from using noisy pseudolabels or perpetuate a model bias. Unsupervised pixel contrastive learning has been used in SSL to encourage compact clusters of representations \citep{AlonsoSSL, kwon2022ELN, liu2022reco}. This mechanism pulls together positive pairs of pixels in the latent space, while pulling negative pairs apart to increase separability. Moreover, supervised pixel contrastive learning has been proposed as a regularizer of the embedding space to encourage better clusterability \citep{WangPCL, pissas2022multi}, boosting the performance of fully supervised methods. \rebuttal{Even if SSL frameworks may share some elements with SSDA methods, they should be properly modified to account for domain adaptation in order to leverage source domain data. The interplay between DA and SSL mechanisms, which we study in this work, is not trivial to predict and requires careful consideration.}

\subsection{Semi-Supervised Domain Adaptation (SSDA)} 
In SSDA the learner has access to source labels, target unlabeled data and a few target labels. SSDA is less explored in the literature, only recently it has received more attention in image classification \citep{Saito2019MME, Berthelot_adamatch, Qin2021uoda, kim2020attract}. While most methods are based on UDA's core idea of domain alignment, \citep{Mishra_PAC} notice that a few target labels are sufficient in SSDA to forego domain alignment and focus on target feature clusterability instead. \rebuttal{However, the dense task of semantic segmentation is more complex than image classification, requiring SSDA methods to be revisited and developed for this setting. Additional challenges of the task are the uncertainty in pixels (e.g., at boundaries between objects), which impedes the use of explicit entropy minimization \citep{Saito2019MME}, and a large class imbalance.}


So far, two frameworks have been devised for SSDA in semantic segmentation~\citep{Wang_ASS, ChenDual}, and one more considers the extension from SSL \citep{AlonsoSSL}. \citet{Wang_ASS} uses adversarial training to align the domains at two representation levels, local and global, but fails to fully leverage the few target labels. \citet{ChenDual} base their method on domain mixing and iterative self-training, with the goal of aligning source and target domain representations.
The domain mixup is achieved with CutMix \citep{yun2019cutmix} and by mixing domains in the mini-batch. Lastly, \citet{AlonsoSSL} propose an SSL framework with consistency regularization and pixel contrastive learning. They also investigate the extension to SSDA by adding source data, but only find a moderate improvement since they do not take domain alignment considerations. 

\section{Method}

\begin{figure}
    \centering
    \includegraphics[width=0.6\linewidth]{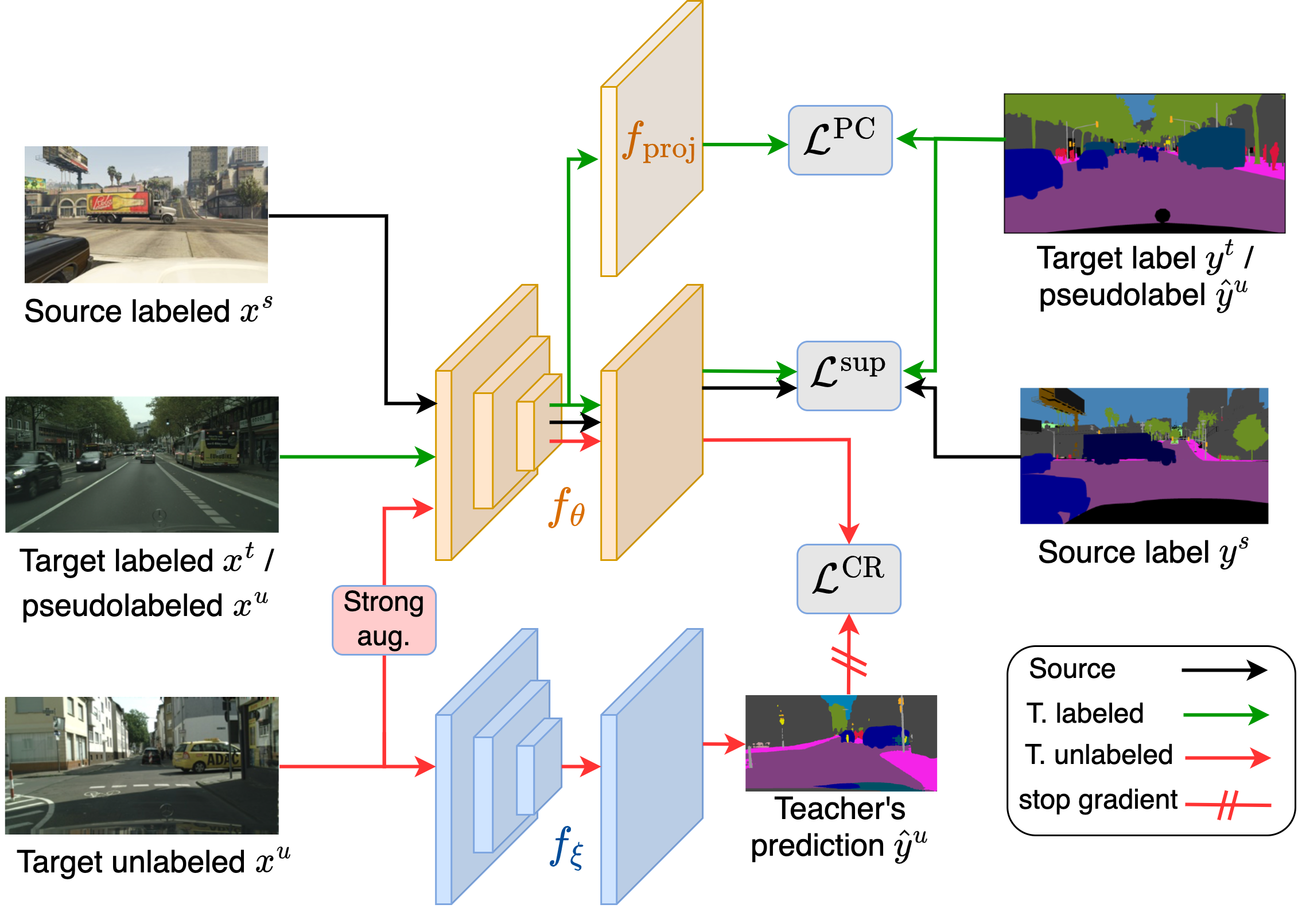}
    \caption{\textbf{Framework overview.} In each round, we train a student model $f_\theta$ with a combination of supervised learning $\mathcal{L}^\textrm{sup}$, consistency regularization (CR) $\mathcal{L}^\textrm{CR}$ and pixel contrastive learning $\mathcal{L}^\textrm{PC}$. We use a mean teacher $f_\xi$ to generate pseudotargets in CR, and stop its gradient.
    In subsequent rounds of self-training, the target labeled set includes pseudolabels generated in the previous round. 
    }
    \label{fig:method_schematic}
\end{figure}

In this section, we present our framework for SSDA semantic segmentation. In SSDA, we have access to a source labeled dataset $\mathcal{D}_s = \{(x_i^s, y_i^s)\}_{i=1}^{N_s}$, a few target labeled samples $\mathcal{D}_{t} = \{(x_i^t, y_i^t)\}_{i=1}^{N_t}$ and a set of target unlabeled samples $\mathcal{D}_{u} = \{x_i^u\}_{i=1}^{N_u}$, where typically $N_t \ll N_u$.

The main goal of our framework is to encourage tight clustering of target representations, such that similar pixels are clustered together in the latent space and the identity of each cluster is inferred from the few labels, a key idea in SSL. 
Moreover, we consider domain alignment to better leverage source data, such that source and target representations are aligned and the model can generalize to both domains.
A schematic of the framework is depicted in Fig. \ref{fig:method_schematic}. We use a student-teacher scheme, keeping a set of parameters $\theta$ for the student model $f_\theta$ and parameters $\xi$ for the teacher model $f_\xi$. The teacher model $f_\xi$ is an exponential moving average (EMA) of $f_\theta$ with coefficient $\mu \in [0,1]$, which provides more robust predictions \citep{MeanTeacher}. The parameters of $f_\xi$ are updated by $\xi = \mu \xi + (1-\mu)\theta$.

In the next subsections we present each of the components of the framework: a supervised objective (Sec. \ref{sec:supervised_training}), consistency regularization (Sec. \ref{sec:consistency_regularization}), pixel contrastive learning (Sec. \ref{sec:pixel_CL}) and an iterative self-training scheme (Sec. \ref{sec:iterative_ST}). Finally, in Sec. \ref{sec:adaptation_to_UDA_and_SSL} we discuss how to extend the framework to the neighboring settings of UDA and SSL.

\subsection{Supervised training on labeled data}
\label{sec:supervised_training}
The available source and target labels are used in a supervised fashion to minimize the cross-entropy with respect to the model predictions. We use class weights to mitigate the class imbalance in semantic segmentation datasets. 
Importantly, we mix source and target batches which helps in learning domain-robust representations \citep{ChenDual}. We define $\mathcal{L}^{\textrm{sup}}$ as
\begin{equation}
    \mathcal{L}^{\textrm{sup}} = \lambda_s \, \mathcal{Q}(f_\theta(x^s), y^s) + \lambda_t \, \mathcal{Q}(f_\theta(x^t), y^t),
    \label{eq:supervised_loss}
\end{equation}
where $\mathcal{Q}(\cdot,\cdot)$ is the weighted cross-entropy. With images of $H \times W$ pixels, one-hot semantic labels as $y$ and $C$  classes, $\mathcal{Q}(\cdot,\cdot)$ is defined as

\begin{equation}
    \mathcal{Q}(\hat{y}, y) = - \frac{1}{H\cdot W} \sum_{j=1}^{H\cdot W} \sum_{c=1}^C \alpha_c \cdot  y_{j,c} \cdot \log(\hat{y}_{j,c}).
\end{equation}
Class weights $\alpha_c$ are computed for $\mathcal{D}_s$ and $\mathcal{D}_t$ separately (see Sec. \ref{tab:HP_configuration}).

\subsection{Consistency Regularization}
\label{sec:consistency_regularization}
Consistency regularization is an unsupervised mechanism that encourages tight and well-separated clusters of representations by promoting consistent predictions between different augmentations of an image. We define $\mathcal{L}^{\mathrm{CR}}$ as the pixel-wise cross-entropy between the prediction of the student $f_\theta$ on a random strong augmentation $x'$ and a one-hot pseudo-target generated by the teacher $f_\xi$ on the original image $x$. The gradient is stopped on the pseudo-target such that $f_\xi$ does not receive any update. The consistency loss for an image $x$ is 
\begin{equation}
    \hat{y}_j = \arg \max f_\xi(x)_j,
\end{equation}
\begin{equation}
    \mathcal{L}^{\textrm{CR}}(x) = \frac{1}{H\cdot W}\sum_{j=1}^{H\cdot W} \textrm{CE}\big( f_\theta(x')_j, \,  \hat{y}_j \big),
    \label{eq:loss_CR_OH}
\end{equation}
where $\textrm{CE}(\cdot, \cdot)$ is the standard cross-entropy loss. This objective leverages unlabeled target data $\mathcal{D}_u$. Details on the transformations used for the random augmentations are provided in Sec. \ref{sec:app_CR_augmentations}.

\subsection{Supervised Pixel Contrastive Learning}
\label{sec:pixel_CL}
To further enhance target feature clusterability, we add a pixel contrastive objective for target labeled data $\mathcal{D}_t$.
With this objecive, pixels of the same class are pushed together in the embedding space, forming more compact clusters, while pixels of different classes are pushed apart, forming low-density regions between clusters. 
A projection head $f_\textrm{proj}$ produces pixel embeddings $z_j$ to be contrasted. The supervised contrastive loss for pixel $j$ is given by


\begin{equation}
    \small\mathcal{L}^\textrm{PC}_j = \frac{1}{|\mathcal{P}_j|}\sum_{z_j^+\in\mathcal{P}_j} - \log \frac{\exp(z_j\cdot z_j^+ / t)}{\exp(z_j\cdot z_j^+ / t) + \sum\limits_{z_j^-\in\mathcal{N}_j} \exp(z_j\cdot z_j^- / t)},
    \label{eq:loss_PCi}
\end{equation}
where $z_j$ is contrasted with a set $\mathcal{P}_j$ of positive samples from the same class and a set $\mathcal{N}_j$ of negative samples from different classes. The symbol $t$ denotes a temperature hyperparameter. A more complex version of this module was introduced by \citet{WangPCL}, but we found the memory bank or the pixel-to-region contrast redundant in our preliminary experiments. Importantly, we apply this objective to target labeled data only, and not unlabeled samples, as relying on ground-truth results in better learnt representations. At each iteration we contrast a subset of pixels sampled from the current batch, up to $N_{\textrm{pix}}$ from each class, using hard example sampling~\citep{WangPCL}. Let $A$ be the total number of pixels sampled from the $\mathcal{D}_t$ batch, with $A \le N_{\textrm{pix}} \cdot C$, then the pixel contrastive loss is defined by

\begin{equation}
    \mathcal{L}^\textrm{PC} = \frac{1}{A} \sum_{j=1}^{A} \mathcal{L}^\textrm{PC}_{j}.
    \label{eq:loss_PC}
\end{equation}

Collecting \eqref{eq:supervised_loss}, \eqref{eq:loss_CR_OH} and \eqref{eq:loss_PC}, the overall loss function to be minimized is given by
\newcommand{\mathsc}[1]{{\footnotesize\textsc{#1}}}
\begin{equation}
    \mathcal{L} = \mathcal{L}^{\textrm{sup}}_{\mathcal{D}_s, \mathcal{D}_t} +  \lambda_1 \, \mathcal{L}^{\mathsc{CR}}_{\mathcal{D}_u} + \lambda_2 \, \mathcal{L}^{\mathsc{PC}}_{\mathcal{D}_t}.
    \label{eq:total_loss}
\end{equation}
We minimize $\mathcal{L}$ in each iteration of a self-training scheme, explained in the next section.

\subsection{Iterative Self-training}
\label{sec:iterative_ST}
In the few-labels regime, the lack of diversity in $\mathcal{D}_t$ is problematic.
To mitigate that we employ an offline self-training algorithm that leverages pseudolabels for unlabeled images in $\mathcal{D}_u$. 
A more diverse pool of labeled samples increases the efficiency of training.
In a second stage of each iteration, we drop the pseudolabels, which innevitably contain some noise, to fine-tune using only ground-truth annotations. The procedure is summarized in Algorithm \ref{alg:ST}, where $\textbf{M}_k$ represents a model trained in the $k^\textrm{th}$ self-training round. \rebuttal{The quality of psuedolabels is critical in self-training. Following \cite{Li2019bidirectional}, we only annotate pixels with a prediction confidence above a threshold $\tau$, and discard the pseudolabel on pixels with uncertain predictions.}



\begin{algorithm}[!h]
\caption{Iterative Self-training}
\label{alg:ST}
    \begin{algorithmic}
    \State Train $\textbf{M}_0$ on $\big[\mathcal{D}_s, \mathcal{D}_t, \mathcal{D}_u \big]$ for $n_\textrm{steps}$. \Comment{First training round}
 
    \State
    \For{$k=\{1, \dots, K\}$} \Comment{Self-training rounds}
        \State $\{\hat{y}_i^u\}_{i=1}^{N_u} = \texttt{generate\_PL}(\mathcal{D}_{u}, \textbf{M}_{k-1})$
        \State $\mathcal{D}_{t+\hat{u}} \gets \mathcal{D}_{t} \, \cup \{(x_i^u, \hat{y}_i^u)\}_{i=1}^{N_u}$ 

        \State Train $\textbf{M}_k$ on $\big[\mathcal{D}_s, \mathcal{D}_{t+\hat{u}}, \mathcal{D}_u \big]$ for steps $[0, n_\textrm{drop})$
        \State Train $\textbf{M}_k$ on  $\big[\mathcal{D}_s, \mathcal{D}_t, \mathcal{D}_u \big]$ for steps $[n_\textrm{drop}, n_\textrm{steps})$
    \EndFor
    \State    
    \State Return $(\textbf{M}_{K-1}, \textbf{M}_K)$. 
    \Comment{Use ensemble at test time}
    \end{algorithmic}
\end{algorithm}

\subsection{Adaptation to UDA and SSL}
\label{sec:adaptation_to_UDA_and_SSL}
In this section we discuss how to adapt our SSDA framework to be used in the UDA and SSL settings. For SSL we simply drop the source data and the supervised loss term becomes $\mathcal{L}^{\textrm{sup}} = \lambda_t \, \mathcal{Q}(f_\theta(x^t), y^t)$. The adaption to UDA has two caveats. Firstly, since we do not have target labeled data, we cannot apply the pixel contrastive learning module on $\mathcal{D}_t$. Therefore, we only use this module on $\mathcal{D}_{t+\hat{u}}$ when pseudolabels are available. Secondly, we modify the consistency regularization formulation to use the teacher's  class probability predictions as pseudo-targets, instead of transforming them into a one-hot encoding. Thus, \eqref{eq:loss_CR_OH} is replaced by
\begin{equation}
    \mathcal{L}^{\textrm{CR}}_\textrm{prob}(x) = \frac{1}{H\cdot W}\sum_{j=1}^{H\cdot W} \textrm{CE}\big( f_\theta(x')_j, \,  f_\xi(x)_j \big).
    \label{eq:loss_CR_PD}
\end{equation}
We observed that using $\mathcal{L}^{\textrm{CR}}_\textrm{prob}$ resulted in more stable training in UDA, while $\mathcal{L}^{\textrm{CR}}$ was stable in SSDA and yielded a slighlty better performance. 

\section{Experiments}
This section presents the experimental setup, SSDA results of the proposed framework, a comparison to UDA and SSL methods, and ablation studies.

\subsection{Implementation Details}
\label{sec:impl_details}

Below we discuss the datasets and model architecture used. As for hyperparamters, we use a fixed training configuration across all experiments and for all datasets, which is detailed in Tab. \ref{tab:HP_configuration} in the Appendix. Experiments are conducted on a single V100 GPU with 32 GB of memory. 

\subsubsection{Datasets}
We use the popular GTA$\rightarrow$Cityscapes as our main semantic segmentation benchmark. Cityscapes, the target dataset, has 2975 training and 500 validation images of European urban scenarios, manually annotated with 19 classes. As standard, we downsample the original resolution of $2048\times1024$ pixels to $1024\times512$ for training. The source GTA dataset \citep{Richter2016gta5} contains 24966 computer-generated urban images for training, which we downsample from $1914\times1052$ to $1280\times720$ pixels, as standard. Labels contain 33 semantic classes, we select only the 19 classes that coincide with Cityscapes, as \cite{Wang_ASS}.

Additionally, we experiment on the datasets of Synthia (source) and BDD (target). Synthia \citep{ros2016synthia} has 9400 synthetic images of $1280\times760$ pixels. It is evaluated on 16 or 13 classes, also present in Cityscapes. For BDD \citep{yu2020bdd100k} we use the 7000 train and 1000 validation real images of US streets at the original resolution of $1280\times720$ pixels. 

For all datasets, we perform random square crops of $512\times512$ and horizontal flips at training time. For evaluation, following standard procedure, we report the mean Intersection over Union (mIoU), averaged over 3 runs with different random labeled/unlabeled training set split.

\subsubsection{Architecture}
We use a DeepLabv2 \citep{Chen2017deeplabv2} decoder and ResNet-101 backbone, for fair comparison with previous works on SSDA \citep{Wang_ASS, ChenDual, AlonsoSSL}, and which is also widely used in UDA benchmarks. The DeepLabv2 decoder uses an ASPP module to obtain multi-scale representations. The ResNet backbone used is always pretrained on ImageNet. Following \cite{WangPCL}, the projection head $f_\textrm{proj}$ for pixel contrast transforms the 2048-dim features from the backbone into 256-dim normalized embeddings. It is composed of two $1\times1$ convolutional layers interleaved with \texttt{ReLU} and \texttt{BatchNorm} layers. 

\subsection{Results}
\subsubsection{SSDA on GTA$\rightarrow$Cityscapes}
\label{sec:results_ssda}
We present our main SSDA results on the widely used GTA$\rightarrow$Cityscapes benchmark in Tab. \ref{tab:SSDA_sota} and Fig. \ref{fig:UDA2SSDA}. We compare our performance to the existing SSDA semantic segmentation methods. Moreover, to provide a competitive baseline, we extend a state-of-the-art UDA method (DAFormer, \cite{Hoyer_DAFormer}) to SSDA. We also include results for training only on labeled target (T) or source and target (S+T) data, and a fully supervised (FS) oracle trained on the entire 2975 target labeled samples. 

Our framework outperforms all previous methods by a substantial margin and sets a new state-of-the-art in the SSDA regime with few labels ($\nicefrac{1}{60}$, $\nicefrac{1}{30}$ and $\nicefrac{1}{15}$ of labeled data). Furthermore, at the most challenging setting of $50$ ($\nicefrac{1}{60}$) target labels, we beat most of the baselines when they use $\times4$ or even $\times10$ more labels. Only when labels are more abundant, at $500$ ($\nicefrac{1}{6}$) target labels, does DAFormer outperform ours, which we speculate is due to the specific measures it takes to generalize in Cityscapes, such as thing-class regularization. 

Compared to supervised performance, with $100$ target labels we already achieve an accuracy of $66.0$ mIoU, only $1.0$ point shy of FS, and surpass it with $200$ of labels. Thus, we demonstrate the potential of SSDA to close the gap to supervised performance at a moderate annotation cost.

\rebuttal{Our method greatly outperforms the previous works tailored to SSDA segmentation. In particular, we do not find necessary to mix domains explicitly as in \cite{ChenDual}, the implicit mixing by using mixed batches (see Sec. \ref{sec:framework_ablation}) and the domain robustness effect of consistency regularization (see Sec. \ref{sec:source_styling}) achieve a better domain alignment.}
We also compare against DAFormer on their Transformer-based architecture (for implementation details see App. \ref{sec:app_other_impl_details}). We find that our method outperforms DAFormer in the semi-supervised low-label regime (Tab.~\ref{tab:UDA_transformer}). However, the gap to supervised performance is still large, interesting future work could be focused on SSDA methods tailored to Transfomers. As \cite{Hoyer_DAFormer} show, this network requires careful design to avoid overfitting to common classes and achieve stable training, which our framework is missing and explains the gap in UDA performance.

\begin{table}[!t]
    \centering
    \caption{GTA$\rightarrow$Cityscapes SSDA semantic segmentation results (mIoU) with a DeepLabv2 + ResNet-101 network. Our framework outperforms baselines in the SSDA low-label regime, achieving near fully supervised (FS) performance at a low annotation cost. All results are averaged over 3 runs.}
    \vspace{3mm}
    \resizebox{0.74\columnwidth}{!}{%
    \begin{threeparttable}
        \begin{tabular}{l| c | c c  c  c |c}
            \toprule
            Target labels & UDA & 50 & 100 & 200 & 500 &  FS \\
            Label ratio & 0 & $\nicefrac{1}{60}$ &  $\nicefrac{1}{30}$ &  $\nicefrac{1}{15}$ &  $\nicefrac{1}{6}$ & 1\\
            \midrule
            $\mathcal{L}^{\textrm{sup}}$ (T) & - & 41.2 & 46.5  & 52.7 & 60.4  & 67.0\\
            $\mathcal{L}^{\textrm{sup}}$ (S+T) & - &52.4\textbf{} & 54.3 &  57.8 & 61.4 & 65.8 \\
            \midrule
            ASS \citep{Wang_ASS}  & - & -& 54.2 & 56.0 &  60.2 &  65.9 \\
            \cite{AlonsoSSL}  & - & - & 59.9 & 62.0  & 64.2  &   67.3\\
            DACS \citep{Tranheden_dacs}~\tnote{*} & 52.1 & - & 61.0 & 63.1 & 64.8 & - \\
            \cite{ChenDual} & - & - & 61.2  & 60.5 & 64.3  & 65.3 \\
            DAFormer \citep{Hoyer_DAFormer}~\tnote{\textdagger} &\textbf{56.0} & 61.8 & 63.5 & 66.3 & \textbf{70.4} & - \\
            Ours & 51.8 & \textbf{64.3} & \textbf{66.0}  & \textbf{67.3}  & 68.3  &  67.0 \\
            \bottomrule
        \end{tabular}
        
        \begin{tablenotes}
        \vspace{2mm}
        \item[*] SSDA results from \cite{Hoyer_SDE}
        \item[\textdagger] on DeepLabv2
        \end{tablenotes}
    \end{threeparttable}
    }
    \label{tab:SSDA_sota}
\end{table}

\begin{table}[!t]
    \centering
    \caption{SSDA semantic segmentation results on GTA$\rightarrow$Cityscapes (mIoU) with a DAFormer network (Transformer-based). We extend DAFormer to the SSDA setting and outperform it at the low-label regime, but fall short of supervised (FS) performance. All results are averaged over 3 runs.}
    \vspace{3mm}
    \begin{threeparttable}
        \resizebox{0.71\columnwidth}{!}{%
        \begin{tabular}{l|c|c c c c}
        \toprule
        Setting & UDA &  \multicolumn{3}{c}{SSDA} \\
        Target labels  &  0  & 50 & 100 & 200 & 500 \\
        Label ratio  &  0  & $\nicefrac{1}{60}$ &  $\nicefrac{1}{30}$ &  $\nicefrac{1}{15}$ &  $\nicefrac{1}{6}$ \\
        \midrule
        \multicolumn{6}{l}{\textit{GTA $\rightarrow$ Cityscapes (DAFormer)} $\,\,$ FS: 77.6 mIoU \citep{Hoyer_DAFormer}} \\
        \midrule 
        DAFormer \citep{Hoyer_DAFormer} & \textbf{68.3} & 66.2 & 69.8 & 71.2 & \textbf{ 74.4}\\
        Ours & 55.5 & \textbf{68.2} & \textbf{71.4} &\textbf{ 72.1} & 73.5 \\
        \bottomrule
        \end{tabular}
        }
    \end{threeparttable}
    \label{tab:UDA_transformer}
\end{table}


\subsubsection{SSDA on other datasets.}
To show the generalization ability of the proposed method to other datasets, we perform experiments on Synthia$\rightarrow$Cityscapes, GTA $\rightarrow$BDD and Synthia$\rightarrow$BDD, all of them $syn2real$ semantic segmentation tasks. We focus on the most challenging SSDA regime, with $\le\nicefrac{1}{30}$ target labels. We use the same training configuration as for GTA$\rightarrow$Cityscapes, without tuning any hyperparameter.

For all datasets, we show that our method is comparable to or outperforms fully supervised (FS) training using only $\nicefrac{1}{30}$ target labels (Tab. \ref{tab:other_datasets}). Moreover, for Synthia$\rightarrow$Cityscapes we also run experiments on DAFormer to provide a competitive baseline, which we beat in all cases. The positive results in other datasets without changing the hyperparameter configuration suggest high practical applicability of the method proposed.

\begin{table}[!t]
    \centering
    \caption{SSDA semantic segmentation results (mIoU) on additional benchmarks. Our method achieves near fully supervised (FS) performance at a low annotation cost. All results are averaged over 3 runs on a DeepLabv2 + ResNet-101 network.}
    \vspace{3mm}
    \resizebox{0.7\columnwidth}{!}{%
    \begin{threeparttable}
        
            \begin{tabular}{l|c | c c}
            \toprule
            \multicolumn{4}{l}{\textit{Synthia$\rightarrow$CS}, 16 (13) classes, FS: 68.9 (73.1) mIoU} \\
            \midrule
            Target labels  &  0  & 50 ($\nicefrac{1}{60}$) & 100 ($\nicefrac{1}{30}$) \\
            $\mathcal{L}^\textrm{sup}$ (S+T) & 29.4 (33.6) & 49.0 (58.4) & 52.5 (61.8)\\
            Ours & - (-)& \textbf{64.5} (\textbf{73.9}) & \textbf{67.2} (\textbf{75.7})\\
            DAFormer \citep{Hoyer_DAFormer}~\tnote{\textdagger} & 53.4 (60.6) & 62.4 (68.0) & 64.6 (70.6)\\
            \midrule
            \midrule
            \multicolumn{4}{l}{\textit{GTA$\rightarrow$BDD}, 19 classes, FS: 55.8 mIoU} \\
            \midrule
            Target labels  &  0  & 100 ($\nicefrac{1}{70}$) & 233 ($\nicefrac{1}{30}$) \\
            $\mathcal{L}^\textrm{sup}$ (S+T) & 33.2 & 48.3 & 51.5 \\
            Ours & 43.1 & \textbf{52.5} & \textbf{54.5} \\
            \midrule
            \multicolumn{4}{l}{\textit{Synthia$\rightarrow$BDD}, 16 classes, FS: 56.6 mIoU} \\
            \midrule 
            $\mathcal{L}^\textrm{sup}$ (S+T) & 24.2 & 43.5 & 48.1 \\
            Ours & - & \textbf{54.5} & \textbf{57.6} \\
            \bottomrule
            \end{tabular}
        
        \vspace{2mm}
        \begin{tablenotes}
        \item[\textdagger] on DeepLabv2.
        \end{tablenotes}
        
    \end{threeparttable}
    }
    \label{tab:other_datasets}
\end{table}

\subsubsection{UDA $\rightarrow$ SSDA.}
\label{sec:uda2ssda}
When no labels are available, our SSDA framework in a UDA setting (see Sec. \ref{sec:adaptation_to_UDA_and_SSL}) achieves $51.8$ mIoU, which is comparable to well-established methods such as DACS \citep{Tranheden_dacs}, but below recent specialized methods. Compared to UDA state-of-the-art, with BAPA \citep{liu2021bapa} achieving an accuracy of $57.4$ mIoU, our method improves by $+6.9$ mIoU using only $50$ target labels. This result demonstrates the high value of even just a few annotations and thus the potential of SSDA. In Sec. \ref{sec:app_UDA_HR} we present an extended comparison to UDA methods, including those using high-resolution images (e.g., HRDA \citep{Hoyer_HRDA}), which we omit here for a fair comparison. 

\begin{figure}[b!]
    \centering
    \includegraphics[width=0.5\linewidth]{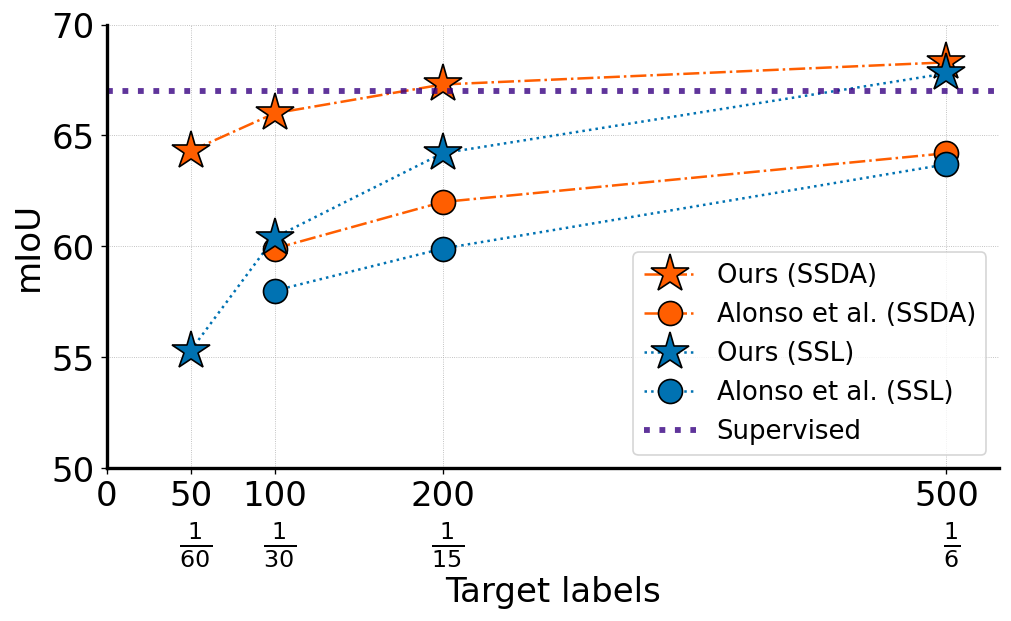}
    \caption{SSL vs. SSDA semantic segmentation results (mIoU) on GTA$\rightarrow$Cityscapes for our method and \cite{AlonsoSSL}. We show a substantial improvement when using source data (SSDA) compared to SSL, particularly in the low-label regime. The difference is less pronounced as more target labels are used. All results are the average of 3 runs on a DeepLabv2 with ResNet-101 backbone.}
    \label{fig:SSL2SSDA}
\end{figure}

\subsubsection{SSL $\rightarrow$ SSDA.}
To quantify the potential improvement of using a source domain, in this section compare our SSDA framework to its SSL counterpart. To our knowledge, only \cite{AlonsoSSL} have compared these settings, which demonstrated only a moderate improvement when using source data. However, we found that when using a framework that takes measures to align domains, adding source domain data can substantially improve performance. In Fig. \ref{fig:SSL2SSDA} we show a direct comparison of SSL vs. SSDA between our method and \cite{AlonsoSSL}. Our method better leverages source data and obtains $+9.0$ mIoU at $50$ labels ($\nicefrac{1}{60}$) and $+5.6$ mIoU at $100$ labels ($\nicefrac{1}{30}$). Interestingly, we observe a trend where the performance boost of SSDA decreases as more target labels are available, shrinking to $+0.5$ mIoU at $500$ labels. We conclude that a source dataset is particularly beneficial when very few target labels are available, as it reduces the risk of SSL to overfit to the few annotations.

\subsection{Ablation studies}
\label{sec:results_ablation}

In this section we explore the impact of each component of the framework. 

\subsubsection{Framework ablation}
\label{sec:framework_ablation}
In Tab. \ref{tab:abl_general} we compare the performance of $\textbf{M}_0$, a model trained on $\mathcal{L}$ (\ref{eq:total_loss}) for one training round, to a number of framework variants. We find that consistency regularization is by far the most important element, as removing $\mathcal{L}^\textrm{CR}$ results in $-8$ mIoU. We also find it important to use class weights to mitigate class imbalance ($-2.3$ mIoU), and to mix source and target data in the same batch in $\mathcal{L}^\textrm{sup}$ ($-0.8$ mIoU), which encourages domain mixing \citep{ChenDual} and helps learn a more domain-robust segmentor. 

Pixel contrastive learning is also found to be a good regularizer, removing $\mathcal{L}^\textrm{PC}$ results in $-1.2$ mIoU (Tab.~\ref{tab:abl_general}). Furthermore, we try two variants of pixel contrastive learning. Firstly, in ``$\mathcal{L}^\textrm{PC}$: +$\mathcal{D}_u$" we adopt the contrastive learning module proposed by \cite{AlonsoSSL}, which uses both labeled and unlabeled data, but observe a performance drop ($-0.5$ mIoU). We attribute the drop to incorrect contrastive pairs on unlabeled pixels, while supervised pixel contrast only relies always on ground-truth. In the second variant, 
``$\mathcal{L}^\textrm{PC}$: +$\mathcal{D}_s$", we try adding source labeled data to pixel contrast, without success ($-0.5$ mIoU). 
Some previous SSDA works even discourage source clusterability \citep{Qin2021uoda}, aiming for source clusters to enclose target representations.

Finally, we report the improvement in performance between the model after the initial round of training ($\textbf{M}_0$) and the final ensemble model after iterative offline self-training ($\textbf{M}_1+\textbf{M}_2$), which brings $+2$ mIoU. 

\definecolor{emerald}{RGB}{0, 180, 180}
\definecolor{redorange}{RGB}{230, 100, 100}
\def\redbar#1{
  {$-#1$ \color{redorange}\rule{\fpeval{#1*0.3}cm}{5pt}   \color{white}\rule{1cm}{5pt}}}
\def\greenbar#1{
  {\color{white}\rule{3cm}{5pt}   \color{emerald}\rule{\fpeval{#1*0.3}cm}{5pt}} $+#1$}
\def\zero{
  {\color{white}\rule{3cm}{5pt}}$0$}

\begin{table}[b!]
    \centering
    \caption{Ablation study of the proposed framework on SSDA GTA$\rightarrow$Cityscapes with 100 target labels ($\nicefrac{1}{30}$). $\Delta$ denotes difference in mIoU to the baseline $\textbf{M}_0$. Experiments are on the initial round of training (i.e., without iterative self-training). We note that consistency regularization is, by far, the most important component. All results are the average of 3 runs on a DeepLabv2 + ResNet-101 architecture.}
    \vspace{3mm}
    \resizebox{0.75\columnwidth}{!}{%
    \begin{tabular}{r|c|l|c}
        \toprule
        \multicolumn{1}{c|}{$\Delta$} & mIoU & \multicolumn{1}{c|}{Configuration} & Steps \\
        \midrule
        \redbar{8} & $56.0$ & No $\mathcal{L}^\textrm{CR}$ & $40$k\\
        \redbar{2.3} & $61.7$ & $\mathcal{L}^\textrm{sup}$: No class weight & $40$k \\
        \redbar{1.2} & $62.8$ & No $\mathcal{L}^\textrm{PC}$ & $40$k\\
        \redbar{0.8} & $63.2$ & $\mathcal{L}^\textrm{sup}$: No batch mix  & $40$k\\
        \redbar{0.5} & $63.5$ & $\mathcal{L}^\textrm{PC}$: +$\mathcal{D}_u$ \citep{AlonsoSSL} & $40$k\\
        \redbar{0.5} & $63.5$ & $\mathcal{L}^\textrm{PC}$: +$\mathcal{D}_s$ & $40$k\\
        \multicolumn{1}{l|}{\zero} & $64.0$  &  $\textbf{M}_0$  & $40$k\\
        \midrule
        \multicolumn{1}{l|}{\greenbar{2}} & $66.0$ & $\textbf{M}_1+\textbf{M}_2$ & $120$k\\
        \bottomrule
    \end{tabular}
    }
    \label{tab:abl_general}
\end{table}

\subsubsection{Iterative self-training}
\begin{figure}[b!]
    \centering
    \includegraphics[width=0.46\linewidth]{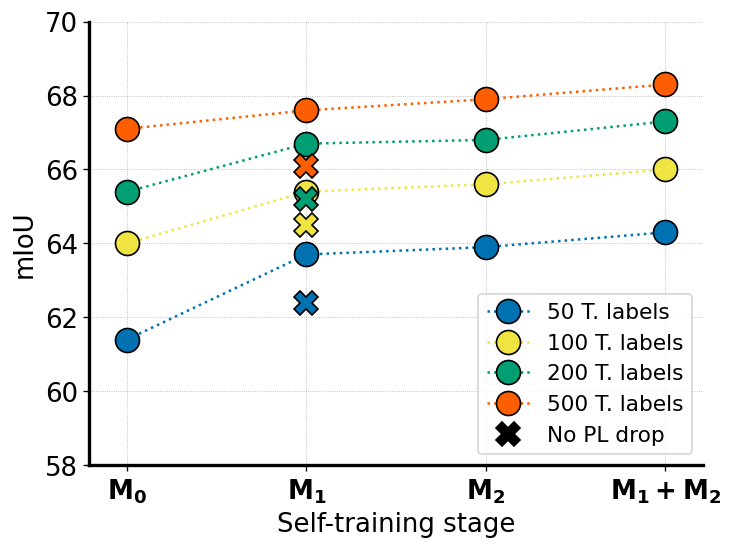}
    \caption{Evolution of performance during self-training from Algorithm \ref{alg:ST}. The first self-training round ($\textbf{M}_0\rightarrow\textbf{M}_1$) brings the largest improvement, the final ensemble ($\textbf{M}_1+\textbf{M}_2$) provides the best performance, and dropping pseudolabels for fine-tuning is beneficial. Results are an average over 3 runs for GTA$\rightarrow$Cityscapes on a DeepLabv2 + ResNet-101 network. A tabular version can be found in Tab. \ref{tab:abl_ST}.}
    \label{fig:abl_ST}
\end{figure}

In Fig. \ref{fig:abl_ST} we break down the impact of self-training. The first round of self-training (between $\textbf{M}_0$ and $\textbf{M}_1$) is the most effective, \rebuttal{while the second round offers marginal to no improvement, indicating convergence of the self-training algorithm.} Finally, the ensemble of $\textbf{M}_1$ and $\textbf{M}_2$ yields the best final performance.

We hypothesize that the main benefit of using pseudolabels is increasing diversity in target samples, which becomes more valuable at low labeling ratios, explaining the larger benefit at 50 target labels. It is also positive to drop pseudolabels after $n_\textrm{drop}$ steps, compared to its counterpart (indicated in Fig. \ref{fig:abl_ST} as ``No PL drop"), and fine-tune on ground-truth annotations.

In Tab. \ref{tab:offline_ST} we compare the self-training scheme to a single longer training round of 120k steps, for the case of $50$ target labels. We find that a self-training scheme was both more effective, as it offered a better final performance ($+0.8$ mIoU), and more efficient, since at 80k steps it already outperformed a single 120k steps training round. The different learning rate decay schedules explains the difference between $\textbf{M}_0$ at 40k steps, the more aggressive decay in self-training allows the model to fine-tune before.

\begin{table}[t!]
    \centering
    \caption{Impact of iterative self-training (rounds of 40k steps) vs. training for longer (one round of 120k steps) on SSDA GTA$\rightarrow$Cityscapes with 50 target labels. Results are the average over 3 runs on a DeepLabv2 + ResNet-101 network.}
    \vspace{3mm}
    \resizebox{0.58\columnwidth}{!}{%
    \begin{threeparttable}
            \begin{tabular}{l | c c | c c}
            \toprule
            Steps & \parbox{2.2cm}{\centering Model \\(Self-training)} & mIoU & \parbox{2.7cm}{\centering Model \\(longer training)} & mIoU\\
            \midrule
            40k &$\textbf{M}_0$ & 61.4 & $\textbf{M}_0$$^*$  & 60.9 \\
            80k &$\textbf{M}_1$  & 63.7 & $\textbf{M}_0$$^*$  & 62.9 \\
            120k &$\textbf{M}_2$ & 63.9 & $\textbf{M}_0$$^*$  & 63.5 \\
            120k &$\textbf{M}_1 + \textbf{M}_2$  & \textbf{64.3} & - & -\\
            \bottomrule
        \end{tabular}
        \begin{tablenotes}
        \item[*] Learning rate decayed linearly during 120k steps
        \end{tablenotes}
    \end{threeparttable}
    }
    \label{tab:offline_ST}
\end{table}

\subsubsection{Source styling and consistency regularization}
\label{sec:source_styling}
Finally, we explored source styling to improve domain alignment. Source styling consists on transforming source images to adopt the target domain style, thus reducing the domain gap in the input space. We tried two transformations, an online normalization of the LAB colorspace \citep{he2021lab} (details in Sec. \ref{sec:app_LAB}) and replacing the original GTA images to GTA stylized as Cityscapes via photorealistic enhancement \citep{richter2022photorealism}. In Tab. \ref{tab:abl_style} we study the interaction of source styling with consistency regularization. When $\mathcal{L}^\textrm{CR}$ is not used, source styling helps, with LAB being most effective. 
Interestingly, source styling did not help when combined with $\mathcal{L}^\textrm{CR}$. 
We hypothesize that, since consistency regularization encourages similar predictions between images under strong augmentations, it may already be promoting a style-invariant model, to a point where styling source data is redundant. On the other hand, artifacts introduced by styling could be harming performance. This observation suggests that consistency regularization is not only promoting compact clustering but also encouraging domain robustness.  

\def\redbar#1{
  {$-#1$ \color{redorange}\rule{\fpeval{#1*0.3}cm}{5pt}   \color{white}\rule{1.2cm}{5pt}}}
\def\greenbar#1{
  {\color{white}\rule{1.1cm}{5pt}   \color{emerald}\rule{\fpeval{#1*0.3}cm}{5pt}} $+#1$}
\def\zero{
  {\color{white}\rule{1.05cm}{5pt}}$0$}
  
\begin{table}[t!]
    \centering
    \caption{Study of the interaction between source styling and consistency regularization (CR). We observe how source styling is beneficial without CR, but harmful when combined with CR, as we hypothesize CR already brings robustness to style. Results are mIoU on the initial round of training $\textbf{M}_0$, an average of 3 runs for experiments on SSDA GTA$\rightarrow$Cityscapes using 100 target labels.}
    \vspace{3mm}
    \resizebox{0.6\columnwidth}{!}{%
    \begin{tabular}{r|c|c l}
        \toprule
        \multicolumn{1}{c|}{$\Delta$} & mIoU & $\mathcal{L}^\textrm{CR}$ & \multicolumn{1}{l}{Source styling} \\
        \midrule
        \multicolumn{1}{l|}{\zero} & $56.0$ & No & No\\
        \multicolumn{1}{l|}{\greenbar{1.4}} & $57.4$ & No &LAB \citep{he2021lab} \\
        \multicolumn{1}{l|}{\greenbar{0.4}} & $56.4$ & No &  photorealistic \citep{richter2022photorealism} \\
        \midrule
        \multicolumn{1}{l|}{\zero} & $64.0$ & Yes & No\\
        \redbar{0.3} & $63.7$ & Yes & LAB \citep{he2021lab} \\
        \redbar{1.2} & $62.8$ & Yes & photorealistic \citep{richter2022photorealism}\\
        \bottomrule
    \end{tabular}
    }
    \label{tab:abl_style}
\end{table}

\section{Discussion}
In this paper, we revisit the SSDA setting in semantic segmentation, which has significant practical implications for industrial and medical imaging applications. We propose a simple SSDA framework that effectively uses the different kinds of data available and achieves fully-supervised accuracy using only a fraction of the target labels. Our method outperforms all SSDA baselines and demonstrates the high value of a handful of target labels to close the gap to supervised performance at a low annotation cost. Our results also demonstrate the generalization ability of the method to other datasets, even without further hyperparameter tuning. 
In addition, we provide insights into several important questions for segmentation practitioners and researchers who aim to minimize annotation costs. These include results on the scalability of existing UDA methods to the semi-supervised setting, as well as a comparison of SSDA and SSL in both low- and high-label regimes. Furthermore, in the following paragraphs, we discuss the relation of SSDA to both UDA and SSL, and propose ways to possibly adapt existing methods to SSDA.

We have demonstrated that existing UDA methods do not perform optimally in the semi-supervised regime, requiring methods tailored to SSDA. To adapt UDA frameworks to SSDA, we propose to consider an objective that emphasizes the tight clustering of target representations, which can be achieved through regularization with supervised pixel contrastive learning. Our findings suggest that domain alignment is less important in SSDA than achieving compact clusters of representations and then identifying them from few-shot samples, as also found by \cite{Mishra_PAC} in image classification. Having demonstrated the potential of SSDA, we encourage future DA research, mostly focused on UDA, to explore SSDA extensions and report results for varying numbers of target labels, in an effort towards a unified learning framework for unlabeled data, similar to \cite{Berthelot_adamatch} for image classification.

Lastly, we observed that SSDA outperforms SSL in the low-label regime, but its advantage diminishes as the number of target labels increases. 
Practitioners facing a performance-vs-cost trade-off may be guided by Fig. \ref{fig:SSL2SSDA} to choose between compiling a source-domain dataset (SSDA) or assuming a larger annotation cost and using SSL.  
Our experiments reveal that to effectively leverage a source dataset, an SSL method must account for domain alignment. In Tab. \ref{tab:abl_general}, we demonstrate that mixing domains in the supervised batch and using exclusively supervised pixel contrast can enhance SSDA performance. 

\section{Acknowledgements}
This work was supported by Hasler Foundation Program: Hasler Responsible AI (project number 21043), by the Army Research Office and was accomplished under Grant Number W911NF-24-1-0048, by the Swiss National Science Foundation (SNSF) under grant number 200021\_205011 and ZEISS Research-IDEAS under grant number 4510852714.

\newpage
\bibliography{main}
\bibliographystyle{tmlr}

\newpage
\appendix
\section{Supplementary material overview}

The supplementary material is organized as follows. We start in Sec. \ref{sec:app_impl_details} with the implementation details. In Sec. \ref{sec:app_CR_augmentations} we detail the augmentations used for consistency regularization. Then we present several additional results, for SSDA in Sec. \ref{sec:app_additional}, for UDA$\rightarrow$SSDA in Sec. \ref{sec:app_UDA_sota}, and for SSL results in Sec. \ref{sec:app_SSL_extended}. We also report per-class performance in Sec. \ref{sec:app_class_accuracy}. In Sec. \ref{sec:app_LAB} we discuss the LAB colorspace transformation used for source styling in the ablation studies, and lastly in Sec. \ref{sec:app_qualitative_res} we show qualitative segmentation results for our framework in the SSDA and UDA settings.

\clearpage
\section{Implementation details}
\label{sec:app_impl_details}

In Tab. \ref{tab:HP_configuration} we list the implementation details used in our trainings to ensure reproducibility of the experiments. \rebuttal{Our algorithm introduces some hyperparameters that should be tuned for each application, we recommend to start with the defaults below, which provided good results in our benchmarks. Perhaps the most important hyperparameters to tune are $\lambda_1$ and $\lambda_2$, the weights for the loss terms. A guideline for their tuning is to pay attention at the loss magnitude of each term, and tune the weights accordingly such that the impact of each component is approximately balanced.}

\begin{table}[!h]
    \centering
    \caption{Implementation details and training hyperparameters.}
    \vspace{3mm}
    \resizebox{0.63\columnwidth}{!}{%
    \begin{tabular}{l|l}
        Training configuration & value \\
        \toprule
        optimizer & SGD, Nesterov momentum $0.9$ \\
        weight decay & $5\times 10^{-4}$ \\
        batch size & $\mathcal{D}_s$: 2, $\,\mathcal{D}_t$: 2, $\,\mathcal{D}_u$: 2  \\
        learning rate & $1\times 10^{-3}$ \\
        learning rate decay & by $\times10$ at $75\%$ of training \\
        $n_\textrm{steps}$ & 40k \\
        $n_\textrm{drop}$ & 20k \\
        self-training rounds &  $K=2$\\
        total steps & 120k (40k$\times3$) \\
        $\lambda_s = \lambda_t = \lambda_1$ & 1 \\
        $\lambda_2$ & 0.2 \\
        gradient clipping norm & 10 \\
        pseudolabels confidence threshold & $\tau=0.9$, following \cite{Li2019bidirectional} \\
        $\mathcal{L}^\textrm{PC}$ warm-up steps & 1k \\
        $N_{\textrm{pix}}$ & 50 \\
        $t$ & 0.1 \\
        EMA decay & $\mu = \min(0.995, \frac{\textrm{step}+1}{\textrm{step}+10})$ \\
        class weight & \parbox{4cm}{$\alpha_c = \sqrt{\frac{f_m}{f_c}}$, following \cite{AlonsoSSL} \\ $f_c$: class frequency \\ $f_m$: median class freq.}\\
        Image augmentations & see Sec. \ref{sec:app_CR_augmentations} \\
    \end{tabular}
    }
    \label{tab:HP_configuration}
\end{table}

\section{Augmentations for Consistency Regularization}
\label{sec:app_CR_augmentations}
The augmentations used for consistency regularization are generated with a series of random transformations, each of them applied with probability $p$. We first apply color jitter ($p=0.8$), Gaussian blur ($p=0.5$) and a modification of RandAugment \citep{Cubuk2020randaugment} ($p=1$) to each image. After that, we apply CutMix \citep{yun2019cutmix}($p=1$) between two images from the batch. 
The modification of RandAugment we use samples from the following subset of augmentations: \texttt{brightness}, \texttt{color}, \texttt{contrast}, \texttt{equalize}, \texttt{posterize}, \texttt{sharpness} and \texttt{solarize}.

\section{Additional SSDA results}
\label{sec:app_additional}

In Tab. \ref{tab:abl_ST} we show the tabular version of Fig. \ref{fig:abl_ST}, which additionally includes the standard deviation in the 3 runs of each experiment.

\begin{table}[h!]
    \centering
    \begin{threeparttable}
        \begin{tabular}{c c|c c c c}
            \toprule
            Model & Steps & 50 ($\frac{1}{60}$) & 100 ($\frac{1}{30}$) & 200 $(\frac{1}{15})$ & 500 ($\frac{1}{6}$) \\
            \midrule
            $\textbf{M}_0$ & 40k & 61.4 \footnotesize{$\pm$ 0.6} & 64.0  \footnotesize{$\pm$ 1.0} & 65.4 \footnotesize{$\pm$ 0.4} & 67.1 \footnotesize{$\pm$ 0.6} \\
            $\textbf{M}_1$ & 80k & 63.7 \footnotesize{$\pm$ 0.9} & 65.4 \footnotesize{$\pm$ 0.8}& 66.7 \footnotesize{$\pm$ 1.0} & 67.6 \footnotesize{$\pm$ 0.2} \\
            $\textbf{M}_2$ & 120k & 63.9 \footnotesize{$\pm$ 1.1}& 65.6 \footnotesize{$\pm$ 0.9}& 66.8 \footnotesize{$\pm$ 0.9}& 67.9 \footnotesize{$\pm$ 0.6} \\
            $\textbf{M}_1+\textbf{M}_2^{\,\,*}$ & 120k & \textbf{64.3} \footnotesize{$\pm$ 0.8}& \textbf{66.0} \footnotesize{$\pm$ 0.5} & \textbf{67.3} \footnotesize{$\pm$ 0.6} & \textbf{68.3} \footnotesize{$\pm$ 0.1} \\
            \midrule
            $\textbf{M}_1^{~\ddagger}$ & 80k & 62.4 \footnotesize{$\pm$ 0.9}& 64.5 \footnotesize{$\pm$ 0.6} & 65.2 \footnotesize{$\pm$ 0.4} & 66.1 \footnotesize{$\pm$ 0.8} \\
            \bottomrule
        \end{tabular}
    \begin{tablenotes}
    \item[*] Ensemble model
    \item[$\ddagger$] Pseudolabels are not dropped after $n_\textrm{drop}$ steps.
    \end{tablenotes}
    \end{threeparttable}
    \caption{Evolution of performance (mIoU) during self-training from \ref{alg:ST}, data corresponding to Fig. \ref{fig:abl_ST}, for different numbers (and ratios) of target labels. The first self-training round ($\textbf{M}_0\rightarrow\textbf{M}_1$) brings the largest improvement, the final ensemble ($\textbf{M}_1+\textbf{M}_2$) provides the best performance, and dropping pseudolabels for fine-tuning is beneficial. Results are the mean and standard deviation over 3 runs for GTA$\rightarrow$Cityscapes on a DeepLabv2 + ResNet-101 network.
    }
    \label{tab:abl_ST}
\end{table}

In Tab. \ref{tab:offline_ST_extended} we show a comparison between using an iterative self-training scheme or training for a single longer round, an extension of Tab. \ref{tab:offline_ST}. In the longer training we use the same hyperparameter configuration except for learning rate decay, which we decay linearly with the number of steps instead of reducing by a factor of $10$ at $75\%$ of iterations. Self-training is both more effective and efficient as we had seen in Tab. \ref{tab:offline_ST}. We also note the trend that the impact of self-training is lower as the number of available target labels increases. Once again, we derive the conclusion that self-training is particularly effective if the number of target labels is very low, as the benefit of increasing the diversity of target annotations becomes larger.

\begin{table}[h!]
    \centering
    \begin{threeparttable}
        \begin{tabular}{l | c c c| c c c}
            \toprule
            Steps & \parbox{2.1cm}{\centering Model \\(Self-training)} & 50 labels & 100 labels & \parbox{2.4cm}{\centering Model \\ (longer training)} & 50 labels & 100 labels\\
            \midrule
            40k &$\textbf{M}_0$ & 61.4 & 64.0 & $\textbf{M}_0^*$  & 60.9 & 63.2 \\
            80k &$\textbf{M}_1$  & 63.7 & 65.4 & $\textbf{M}_0^*$  & 62.9 & 64.8 \\
            120k &$\textbf{M}_2$ & 63.9 & 65.6& $\textbf{M}_0^*$  & 63.5 & 65.5 \\
            120k &$\textbf{M}_1 + \textbf{M}_2$  & 64.3 & 66.0 & - & - & -\\
            \bottomrule
        \end{tabular}
        \begin{tablenotes}
        \item[*] Learning rate decayed linearly during 120k steps
        \end{tablenotes}
    \end{threeparttable}
    \caption{Extended Tab. \ref{tab:offline_ST} that additionally includes results for 100 target labels. Impact of iterative self-training (rounds of 40k steps) vs. training for longer (one round of 120k steps) on SSDA GTA$\rightarrow$Cityscapes. We observe a higher benefit of self-training for the lower number of target labels. Results are the average over 3 runs on a DeepLabv2 + ResNet-101 network.}
    \label{tab:offline_ST_extended}
\end{table}

\clearpage
\section{Additional UDA $\rightarrow$ SSDA results}
\label{sec:app_UDA_sota}

\subsection{UDA on high-resolution images}
\label{sec:app_UDA_HR}
Lately, it has become a trend in UDA to evaluate the proposed methods in a slightly different setting: using full resolution images. The common practice in the well-established GTA$\rightarrow$Cityscapes benchmark has been to downscale the resolution of images (in GTA from $1914\times1052$ to $1280\times720$, in Cityscapes from $2048\times1024$ to $1024\times512$ pixels), to deal with memory constraints of GPUs.  Moreover, downscaled images are also a good benchmark for other datasets in industrial or medical applications which do not count on high-resolution (HR) images. We also note that training on HR images is far less agile, scaling up computational requirements and often requiring multiple GPUs.

Recently, multiple methods have reported results on models trained in original resolution images, which often translate into higher accuracy scores due to the more detailed images. Perhaps surprisingly, this phenomenon has been unnoticed and is not underlined in the literature. We would like to highlight the difference between training on downscaled and HR images and argue that these two settings \textit{should not be directly compared}. The literature in UDA usually presents methods on downscaled and full resolution images indistinctly, which can be misleading when assessing the performance of the algorithm.

We train again our proposed framework on SSDA with 100 labels and obtain $3.3$ mIoU points of improvement, from $66.0$ to $69.3$. In Tab. \ref{tab:UDA_sota_extended} we report an extended comparison to UDA methods, including those that train on HR images. HRDA \citep{Hoyer_HRDA} is the only of such methods that can be trained in a single GPU as they use small a multi-scale scheme and only take small crops of HR images. As the authors demonstrate, HRDA can be used as a plug-in extension to existing UDA methods to further increase performance. We note that SSDA also substantially outperforms UDA on HR images, from $63.0$ mIoU of the state-of-the-art HRDA framework to our $69.3$ mIoU $+6.3$ when adding 100 target labels. We also try our framework in the UDA HR setting, but obtain unstable training, likely missing further regularization to not overfit to the common classes.

\begin{table}[h!]
    \centering
    \begin{threeparttable}
        \begin{tabular}{l|c|c c c c}
        \toprule
        Setting & UDA &  \multicolumn{4}{c}{SSDA} \\
        Target labels  &  0  & 50 & 100 & 200 & 500 \\
        \midrule
        \multicolumn{5}{l}{\textit{Fully supervised ($1024\times512$ pixels):} 67.0 mIoU} \\
        \midrule 
        
        CBST \cite{zou2018cbst} & 45.9 & - & - & - & -\\
        FDA \cite{Yang_FDA} & 50.5 & - & - & - & - \\
        Ours & 51.8 & 64.3 & 66.0 & 67.3 & 68.3 \\
        Ours\tnote{\textdagger} & - & - & 69.3 & - & -\\
        DACS \cite{Tranheden_dacs, Hoyer_SDE} & 52.1 & - & 61.0 & 63.1 & 64.8 \\
        SAC \cite{araslanov2021self} & 53.8 & - & - & - & - \\
        DAFormer \cite{Hoyer_DAFormer, Hoyer_HRDA} & 56.0 & 61.8 & 63.5 & 66.3 & 70.4 \\
        CorDA\tnote{*}& 56.6 & - & - & - & -\\
        BAPA \cite{liu2021bapa}& 57.4 & - & - & - & - \\
        ProDA\tnote{\textdagger} ~~\cite{Zhang_proda} & 57.5 & - & - & - & - \\
        EHTDI\tnote{\textdagger} ~~\cite{li2022ehtdi} & 58.8 & - & - & - & - \\
        CPSL\tnote{\textdagger} ~~\cite{li2022cpsl} & 60.8 & - & - & - & - \\
        DBB\tnote{\textdagger} ~~\cite{chen2022dbb} & 62.7 & - & - & - & - \\
        HRDA\tnote{\textdagger} ~~\cite{Hoyer_HRDA} & \textbf{63.0} & - & - & - & - \\
        \bottomrule
        \end{tabular}
        \begin{tablenotes}
        \item[*] Additional information used for depth-estimation.
        \item[\textdagger] Use of HR images (Cityscapes: $2048\times1024$).
        \end{tablenotes}
    \end{threeparttable}
    \caption{Comparison of SSDA to UDA state-of-the-art methods for semantic segmentation on GTA$\rightarrow$Cityscapes (mIoU), including methods using HR images. Our method improves $+3.3$ mIoU when training on HR images, and outperform the state-of-the-art UDA framework (HRDA \cite{Hoyer_HRDA}) by $6.3$ points. All results are averaged over 3 runs on a DeepLabv2 + ResNet-101 network.}
    \label{tab:UDA_sota_extended}
\end{table}

Likewise, with Transformer-based architectures the use of HR images can also boost performance. HRDA \citep{Hoyer_HRDA} shows an improvement over DAFormer (see Tab. \ref{tab:UDA_transformer_HR}) when using their multi-scale module that leverages high resolution. They outperform our SSDA results on the downscaled Cityscapes, since, as discussed in Sec. \ref{sec:results_ssda}, our framework is not designed for Transformer architectures. Future work could address the design of an SSDA method tailored to Transformers, which requires careful design in order to avoid overfitting and achieve stable training.

\begin{table}[h!]
    \centering
    \begin{threeparttable}
        \begin{tabular}{l|c|c c c c}
        \toprule
        Setting & UDA &  \multicolumn{3}{c}{SSDA} \\
        Target labels  &  0  & 50 & 100 & 200 & 500 \\
        Label ratio  &  0  & $\nicefrac{1}{60}$ &  $\nicefrac{1}{30}$ &  $\nicefrac{1}{15}$ &  $\nicefrac{1}{6}$ \\
        \midrule
        \multicolumn{6}{l}{\textit{Fully supervised: 77.6 \cite{Hoyer_DAFormer}}} \\
        \midrule 
        Ours & 55.5 & 68.2& 71.4 & 72.1 & 73.5 \\
        DAFormer \cite{Hoyer_DAFormer} & 68.3 & 66.2 & 69.8 & 71.2 & 74.4\\
        HRDA\tnote{\textdagger} ~~\cite{Hoyer_HRDA}& 73.8 & - & - & - & -\\
        \bottomrule
        \end{tabular}
        \begin{tablenotes}
        \item[\textdagger] Use of HR images (Cityscapes: $2048\times1024$).
        \end{tablenotes}
    \end{threeparttable}
    \caption{UDA and SSDA semantic segmentation on GTA$\rightarrow$Cityscapes (mIoU) on a Transformer-based architecture (MiT-B5 backone + DAFormer decoder). We extend our results to the Transformer architecture without further adapting the learning algorithm, which explains the lower performance in UDA and the still large gap between SSDA and supervised performance. We also extend DAFormer's \citep{Hoyer_DAFormer} experiments to SSDA. All results are averaged over 3 runs on a DAFormer network. }
    \label{tab:UDA_transformer_HR}
\end{table}

\subsection{Other Implementation Details}
\label{sec:app_other_impl_details}
\textbf{Extending UDA methods to SSDA.} In Fig. \ref{fig:UDA2SSDA} we compare our method to the extension of existing UDA methods to SSDA. For DACS \citep{Tranheden_dacs}, we take the results from \cite{Hoyer_SDE}, which already investigates the extension of this method to SSDA to provide a competitive baseline. For DAFormer \citep{Hoyer_DAFormer}, SSDA results are our own. We base our implementation on the codebase provided by \cite{Hoyer_DAFormer} and keep the exact same hyperparameters for training. We make only two changes. Firstly, we replace the Transformer-based architecture with a DeepLabv2 + ResNet101. Secondly, to adapt to SSDA, we add a loss term with cross-entropy on target labeled data, similar to their cross-entropy loss on source labeled data. For experiments in Tab. \ref{tab:UDA_transformer} we keep the original Transformer architecture, so only the second change applies.

\textbf{High resolution.} In our experiments with high resolution Tab. \ref{tab:UDA_sota_extended} we do not downsample images, keeping Cityscapes at $2048\times1024$ pixels and GTA at $1914\times1052$. During training, to fit into the memory constraints of a single GPU, we make random crops of $756\times756$ pixels, which combined with the small batch size of 2 allow to respect memory constraints. 

\textbf{Training details on DAFormer architecture.} We slightly modify the hyperparameter configuration when training our framework on a DAFormer architecture. For a fair comparison to \cite{Hoyer_DAFormer}, we reduce the number of training steps in order to match their $40$k total iterations. We take $20$k iterations per training round and only perform one round of self-training ($K=1$), such that the total amount of steps is $40$k. The rest of the hyperparameters remain as in Sec. \ref{sec:impl_details}. We ensemble the models of the two rounds for the final result. As for the adaption of the architecture, we add a projection head for pixel contrastive learning which takes the 256-dim output from the feature extractor and projects it through two $1\times1$ convolutional layers interleaved with \texttt{ReLU} and \texttt{BatchNorm} layers, into a new 256-dim embedding.

\clearpage
\section{Additional SSL results}
\label{sec:app_SSL_extended}
In this section, we present additional results regarding experiments in the SSL setting. In Tab. \ref{tab:SSL_sota} we present a comparison to other SSL methods that use the same network as us, a DeepLabv2 + ResNet101 network. 
Our SSDA method performs well when applied to SSL, with comparable or superior accuracy than previous works on the same architecture. Nonetheless, it is worth noting that recent methods using more capable architectures such as a DeepLabv3+ encoder, such as \cite{chen2021CPS, fan2022ucc}, which we do not compare to here, can attain higher performance.

\begin{table}[h!]
    \centering
    \begin{threeparttable}
        \begin{tabular}{l|c c c c|c}
            \toprule
            Labels (ratio) & 50 ($\nicefrac{1}{60}$) & 100 ($\nicefrac{1}{30}$) & 372 $(\nicefrac{1}{8})$ & 744 ($\nicefrac{1}{4}$) & FS \\
            \midrule
            French et al. \cite{french2019cutmix} & - & 51.2 & 60.3 & 63.9 & 67.5 \\
            ClassMix \cite{olsson2021classmix} & - & 54.1 & 61.3 & 63.6 &  66.2 \\
            GuidedMix-Net \cite{Tu2022guidedmix} & - & 56.9 & 65.8 & 67.5 & - \\
            Alonso et al. \cite{AlonsoSSL} & - & 59.4 & 64.4 & 65.9 & 67.3 \\
            Ours & 55.3 & 60.4 & 66.5 & 67.2 & 67.0 \\
            \bottomrule
        \end{tabular}
    \end{threeparttable}
    \caption{Comparison of SSL semantic segmentation methods on GTA$\rightarrow$Cityscapes on DeepLabv2 + ResNet-101 backbone. Our SSDA method applied to the SSL setting performs well, matching or beating other methods for all target label ratios. We only compare here to methods that report results using the same DeepLabv2 network. All results (mIoU) are the average of 3 runs.}
    \label{tab:SSL_sota}
\end{table}

In Tab. \ref{tab:SSL_2_SSDA} we present the results for SSL$\rightarrow$SSDA corresponding to Fig. \ref{fig:SSL2SSDA}, a direct comparison between our method and \cite{AlonsoSSL}, the only previous SSL framework that included SSDA in their experiments. Our method manages to better leverage the source domain, showing an improvement when adding source data, particularly when very few target labels are available.

\begin{table}[h!]
    \centering
    \begin{threeparttable}
        \begin{tabular}{l|c c c c}
            \toprule
            Target labels &50 ($\nicefrac{1}{60}$) &100 ($\nicefrac{1}{30}$) & 200 $(\nicefrac{1}{15})$ & 500 ($\nicefrac{1}{6}$)\\
            \midrule
            Alonso et al.\tnote{*}~~ \cite{AlonsoSSL} & - &58.0 & 59.9 & 63.7 \\
            $\quad$ + source & - &59.9 (+1.9) & 62.0 (+2.1) & 64.2 (+0.5) \\
            \midrule
            Ours (SSL) & 55.3 & 60.4 & 64.2 &  67.8 \\
            $\quad$ + source  & 64.3 (+9.0) & 66.0 (+5.6) & 67.3 (+3.1) & 68.3 (+0.5)\\
            \bottomrule
        \end{tabular}
        \begin{tablenotes}
        \item[*] {\small ImageNet pretrained, lower accuracy than in Tab. \ref{tab:SSL_sota}, which was COCO pretrained.}
        \end{tablenotes}
    \end{threeparttable}
    \caption{Results corresponding to Fig. \ref{fig:SSL2SSDA}. Comparison of SSL$\rightarrow$SSDA semantic segmentation methods on GTA$\rightarrow$Cityscapes. Our proposed framework shows a larger improvement when adding source data. We also note that SSDA outperforms SSL especially when few target labels are available, while at $500$ the improvement is marginal. All results (mIoU) are the average of 3 runs on a DeepLabv2 with ResNet-101 backbone.}
    \label{tab:SSL_2_SSDA}
\end{table}

In Tab. \ref{tab:SSL_ST} we present an ablation of the impact of training rounds during the iterative offline self-training scheme in SSL. Interestingly, we observe how the pseudolabels help when only a few labels are available, possibly explaining the improvement in SSL accuracy over a similar method such as \cite{AlonsoSSL}, which does not use self-training. On the other hand, at a higher labeling ratio the scheme is no longer helpful. This leads us to hypothesize that pseudolabels are especially effective under few labels as they increase the diversity of target labels, while the impact is reduced in the presence of more ground-truth labels. In the latter case, the noise or bias in pseudolabels can even be counterproductive.

\begin{table}[h!]
    \centering
    \begin{threeparttable}
        \begin{tabular}{c c|cccc}
            \toprule
            Model & Steps & 50 ($\nicefrac{1}{60}$) & 100 ($\nicefrac{1}{30}$) & 372 $(\nicefrac{1}{8})$ & 744 ($\nicefrac{1}{4}$) \\
            \midrule
            $\textbf{M}_0$ & 40k & 52.0 \footnotesize{$\pm$ 2.9} & 57.3 \footnotesize{$\pm$ 0.9} & 64.8 \footnotesize{$\pm$ 1.0} & 66.9 \footnotesize{$\pm$ 0.8} \\
            $\textbf{M}_1$ & 80k & 54.7 \footnotesize{$\pm$ 3.2} & 59.6 \footnotesize{$\pm$ 1.0} & 66.0 \footnotesize{$\pm$ 0.5} & 66.8 \footnotesize{$\pm$ 0.8} \\
            $\textbf{M}_2$ & 120k & 55.4 \footnotesize{$\pm$ 3.2} & 60.3 \footnotesize{$\pm$ 1.4} & 66.0 \footnotesize{$\pm$ 0.3} & 66.5 \footnotesize{$\pm$ 0.8} \\
            $\textbf{M}_1+\textbf{M}_2^{\,\,*}$ & 120k & 55.3 \footnotesize{$\pm$ 7.1} & \textbf{60.4} \footnotesize{$\pm$ 1.1} & \textbf{66.5} \footnotesize{$\pm$ 0.1} & \textbf{67.2} \footnotesize{$\pm$ 0.5} \\
            \bottomrule
        \end{tabular}
    \begin{tablenotes}
    \item[*] Ensemble model
    \end{tablenotes}
    \end{threeparttable}
    \caption{Impact of offline self-training in the proposed framework on an SSL setting. Accuracy (mIoU) on Cityscapes validation set on 100 target labels, all results are an average of 3 runs on a DeepLabv2 + ResNet-101 network.}
    \label{tab:SSL_ST}
\end{table}

Lastly, in Tab. \ref{tab:abl_SSL} we present two more ablation results in the SSL setting for 100 labels. We are interested in investigating the impact of using our pixel contrastive learning module (supervised) or an alternative pixel contrastive learning such as the module presented in \cite{AlonsoSSL} (supervised + unsupervised), which we refer to as $\mathcal{L}^\textrm{PC}$: +$\mathcal{D}_u$. While in SSDA the latter performed worse (Tab. \ref{tab:abl_general}), in SSL we obtain the same performance as the baseline $\textbf{M}_0$. We hypothesize that adding unlabeled data to contrastive learning helps in SSL as it avoids relying too much on the few target labels, which we risk overfitting to. Meanwhile, in SSDA this benefit is less pronounced, as the presence of source labels already prevents overfitting, and the inherent noise of unlabeled contrastive learning (due to wrongfully assigned pairs) actually harms performance. In Tab. \ref{tab:abl_SSL} we also confirm that the self-training scheme is beneficial in the SSL setting.

\def\redbar#1{
  {$-#1$ \color{redorange}\rule{\fpeval{#1*0.3}cm}{5pt}   \color{white}\rule{1.2cm}{5pt}}}
\def\greenbar#1{
  {\color{white}\rule{1.5cm}{5pt}   \color{emerald}\rule{\fpeval{#1*0.3}cm}{5pt}} $+#1$}
\def\zero{
  {\color{white}\rule{1.5cm}{5pt}}$0$}

\begin{table}[h!]
    \centering
    \begin{tabular}{l|c|l|c}
        \toprule
        \multicolumn{1}{c|}{$\Delta$} & mIoU & \multicolumn{1}{c|}{Configuration} & Steps \\
        \midrule
        \zero & 57.3 & $\mathcal{L}^\textrm{PC}$: +$\mathcal{D}_u$ \cite{AlonsoSSL} & 40k\\
        \zero & 57.3  &  $\textbf{M}_0$ & 40k\\
        \midrule
        \greenbar{3.1} & 60.4 & $\textbf{M}_1 + \textbf{M}_2$ & 120k\\
        \bottomrule
    \end{tabular}
    \caption{Ablation study of the proposed framework on an SSL setting. Accuracy (mIoU) on Cityscapes validation set on 100 target labels, all results are an average of 3 runs on a DeepLabv2 + ResNet-101 network.}
    \label{tab:abl_SSL}
\end{table}

\clearpage
\newpage
\section{Per-class performance}
\label{sec:app_class_accuracy}
In this section we break down the performance of our framework to per-class performance. We report the Intersection over the Union (IoU) for the 19 semantic classes used in Cityscapes. We start analyzing in Tab. \ref{tab:UDA_class_split} the difference between using $\mathcal{L}^{\textrm{CR}}$ and $\mathcal{L}^{\textrm{CR}}_\textrm{prob}$ for the UDA version of our framework, which we presented in Sec. \ref{sec:adaptation_to_UDA_and_SSL}. We observe how the variant that uses $\mathcal{L}^{\textrm{CR}}_\textrm{prob}$ performs much better than its counterpart, particularly for a set of classes highlighted in bold font. What these classes have in common is that they are underrepresented in the training set, as they are either rare or correspond to small objects. Using the class probabilities in consistency regularization seems to mitigate the issue of overfitting to the most common classes. Interestingly, we do not observe the same behavior in SSDA, with $\mathcal{L}^{\textrm{CR}}$ yielding slightly better performance than $\mathcal{L}^{\textrm{CR}}_\textrm{prob}$. We hypothesize that the presence of a few labels helps the model learn the rare classes, which then are also used as one-hot pseudo-targets in $\mathcal{L}^{\textrm{CR}}$.

Also in Tab. \ref{tab:UDA_class_split} we report per-class performance of our method on SSDA with 50 and 100 target labels. We observe an improvement in all classes compared to UDA, particularly large for uncommon classes (e.g., terrain, sidewalk, train).

\begin{table*}[h!]
    \centering
    \setlength\tabcolsep{1.5pt}
    \renewcommand{\arraystretch}{1.2}
    \resizebox{\textwidth}{!}{%
    \begin{tabular}{|ll|ccccccccccccccccccc|c|}
        \toprule
        Method~~ & Setting  & \rotatebox{90}{road} & \rotatebox{90}{sidewalk} & \rotatebox{90}{building} & \rotatebox{90}{wall} & \rotatebox{90}{fence} & \rotatebox{90}{pole} & \rotatebox{90}{light} & \rotatebox{90}{sign} & \rotatebox{90}{veg.} & \rotatebox{90}{terrain} & \rotatebox{90}{sky} & \rotatebox{90}{person} & \rotatebox{90}{rider} & \rotatebox{90}{car} & \rotatebox{90}{truck} & \rotatebox{90}{bus} & \rotatebox{90}{train} & \rotatebox{90}{motorbike}  & \rotatebox{90}{bike} & mean IoU\\
        \midrule 
        Ours &UDA, $\mathcal{L}^{\textrm{CR}}$ & 94.2 & 0.0 & 82.6 & 3.9 & 3.9 & 3.5 & 41.5 & 47.6 & 84.8 & 11.8 & 87.7 & 46.0 & 3.3 & 87.9 & 59.6 & 49.7 & 0.0 & 34.1 & 0.2 & 39.0 \\
        Ours & UDA, $\mathcal{L}^{\textrm{CR}}_\textrm{prob}$ & 91.9 & \textbf{42.0} & 86.5 & \textbf{26.7} & \textbf{32.2} & \textbf{35.7} & 43.0 & 54.4 & 82.8 & 18.0 & 80.1 & \textbf{65.3} & \textbf{37.3} & 88.7 & 54.9 & 53.7 & 0.0 & 36.4 &\textbf{ 54.5} & 51.8 \\
        \midrule
        Ours &SSDA (50) & 96.6 & 75.4 & 88.8 & 50.4 & 46.4 & 44.9 & 51.6 & 60.9 & 88.6 & 50.1 & 90.9 & 69.5 & 47.2 & 91.7 & 74.5 & 64.8 & 19.6 & 45.3 & 64.9 & 64.3\\
        Ours &SSDA (100) & 96.8 & 76.6 & 89.1 & 51.5 & 46.7 & 46.1 & 51.9 & 62.4 & 89.0 & 51.7 & 91.2 & 69.8 & 48.6 & 91.9 & 74.2 & 69.5 & 35.9 & 47.4 & 64.9 & 66.0 \\

        \bottomrule
    \end{tabular}
    }
    \renewcommand{\arraystretch}{1}
    \caption{Performance per class of our proposed method the UDA and SSDA settings. Average IoU (in \%) over 3 seeds on the 19 Cityscapes classes using a \textbf{DeepLabv2 + ResNet-101 network}. We compare UDA using a one-hot encoding for consistency regularization pseudo-targets ($\mathcal{L}^{\textrm{CR}}$) or the predicted class probability by the teacher ($\mathcal{L}^{\textrm{CR}}_\textrm{prob}$). In \textbf{bold} we highlight the classes with a substantial difference in performance. Finally, we also compare to our framework on SSDA with 50 and 100 target labels, which uses $\mathcal{L}^{\textrm{CR}}$.}
    \label{tab:UDA_class_split}
\end{table*}

In Tab. \ref{tab:class_split} we present the class performance for our method and DAFormer using the Transformer-based architecture presented in \cite{Hoyer_DAFormer}. Our method does not perform well in UDA compared to DAFormer, as it overfits to common classes and performs worse on rare classes (e.g., sidewalk, train). As shown in \cite{Hoyer_DAFormer}, using the MiT-b5 Transformer as backbone requires of specific measures to stabilize training and avoid overfitting to the common classes, which our framework does not have. However, in SSDA with 100 target labels our method improves and is able to learn reasonably well, including the rare classes, ultimately outperforming DAFormer at the same number of labels. This observation suggests that adding a few labels is an alternative way to mitigate overfitting issues in UDA with Transformers, as opposed to using additional regularization terms and algorithmic solutions tailored to a particular dataset.

\begin{table*}[h!]
    \centering
    \setlength\tabcolsep{1.5pt}
    \renewcommand{\arraystretch}{1.2}
    \resizebox{\textwidth}{!}{%
    \begin{tabular}{|ll|ccccccccccccccccccc|c|}
        \toprule
        Method & Setting & \rotatebox{90}{road} & \rotatebox{90}{sidewalk} & \rotatebox{90}{building} & \rotatebox{90}{wall} & \rotatebox{90}{fence} & \rotatebox{90}{pole} & \rotatebox{90}{light} & \rotatebox{90}{sign} & \rotatebox{90}{veg.} & \rotatebox{90}{terrain} & \rotatebox{90}{sky} & \rotatebox{90}{person} & \rotatebox{90}{rider} & \rotatebox{90}{car} & \rotatebox{90}{truck} & \rotatebox{90}{bus} & \rotatebox{90}{train} & \rotatebox{90}{motorbike}  & \rotatebox{90}{bike} & mean IoU\\

        \midrule 
        Ours & UDA & 80.2 & 26.6 & 87.4 & 49.5 & 39.2 & 40.2 & 48.2 & 41.4 & 87.5 & 37.3 & 84.9 & 63.3 & 20.9 & 90.9 & 63.7 & 66.0 & 23.4 & 36.0 & 48.5 & 54.4 \\
        DAFormer & UDA & 95.7 & 70.2 & 89.4 & 53.5 & 48.1 & 49.6 & 55.8 & 59.4 & 89.9 & 47.9 & 92.5 & 72.2 & 44.7 & 92.3 & 74.5 & 78.2 & 65.1 & 55.9 & 61.8 & 68.3 \\
        \midrule
        Ours & SSDA (100)  & 97.2 & 78.5 & 90.2 & 54.8 & 45.7 & 51.1 & 60.0 & 68.2 & 90.9 & 57.3 & 93.7 & 73.0 & 50.8 & 92.7 & 76.8 & 80.4 & 69.7 & 57.0 & 69.7 & 71.4 \\
        DAFormer & SSDA (100) & 97.3 & 78.4 & 89.7 & 42.2 & 44.6 & 50.9 & 58.7 & 66.0 & 90.3 & 58.2 & 93.1 & 74.4 & 49.4 & 92.4 & 68.8 & 79.7 & 71.7 & 50.7 & 68.8 & 69.8\\
        \bottomrule
    \end{tabular}
    }
    \renewcommand{\arraystretch}{1}
    \caption{Performance per class of our proposed method in the UDA and SSDA settings. Average IoU (in \%) over 3 seeds on the 19 Cityscapes classes on a\textbf{ DAFormer network }\cite{Hoyer_DAFormer}. We observe how our method does not perform well in UDA compared to DAFormer, as it overfits to common classes and performs worse on rare classes (e.g., sidewalk, train). However, in SSDA with 100 target labels our method improves and is able to learn reasonably well also the rare classes, ultimately outperforming DAFormer at the same number of labels.}
    \label{tab:class_split}
\end{table*}

\clearpage
\section{LAB colorspace transformation}
\label{sec:app_LAB}

In Sec. \ref{sec:results_ablation} we perform an ablation study investigating the interaction of consistency regularization with source styling. We try two transformations of source images into target domain, namely a LAB colorspace transformation \cite{he2021lab} and photorealistic enhancement \citep{richter2022photorealism}. The latter is of straightforward application; the original source dataset of GTA is replaced by a version of GTA stylized as with Cityscapes style, a dataset which is provided by \cite{richter2022photorealism}. On the contrary, the LAB colorspace transformation is applied online during training. Implemented following \cite{he2021lab}, it consists on matching the statistics of each LAB color channel of a source image to those of a target domain image. Letting $x_{\textrm{LAB}}$ be the image in a LAB colorspace and $\mu$ and $\sigma$ the mean and variance respectively of each channel in this colorspace, then the normalization of a source image with target domain statistics is applied as
\begin{equation}
    \hat{x}^s_{\textrm{LAB}} = \frac{(x^s_{\textrm{LAB}} - \mu_s)}{\sigma_s} \cdot \sigma_t + \mu_t.
    \newline
    \label{eq:lab}
\end{equation}

\begin{figure}[h!]
    \centering
    \includegraphics[width=0.6\linewidth]{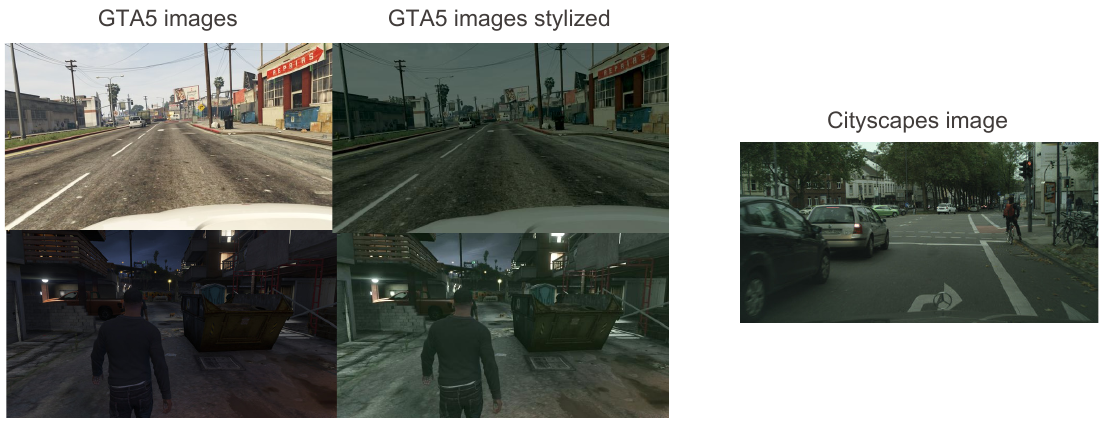}
    \caption{Example of GTA images (source) stylized as a Cityscapes images (target) using LAB colorspace transformation \citep{he2021lab}.}
    \label{fig:LAB}
\end{figure}

In each training step, source and target images are transformed from RGB to LAB colorspace, then the color channels of source images are normalized to the values of a random target image in the batch, following \ref{eq:lab}, and lastly images are transformed back to the RGB space. Examples of GTA (source) images stylized as a Cityscapes (target) sample are shown in Fig.~\ref{fig:LAB}. We hypothesize that the better results with LAB styling are partly due to this online application, as source images are stylized differently at each iteration, possibly leading into a more robust model. 

\clearpage
\section{Qualitative segmentation results}
\label{sec:app_qualitative_res}
\begin{figure}[h!]
    \centering
    \includegraphics[width=\columnwidth]{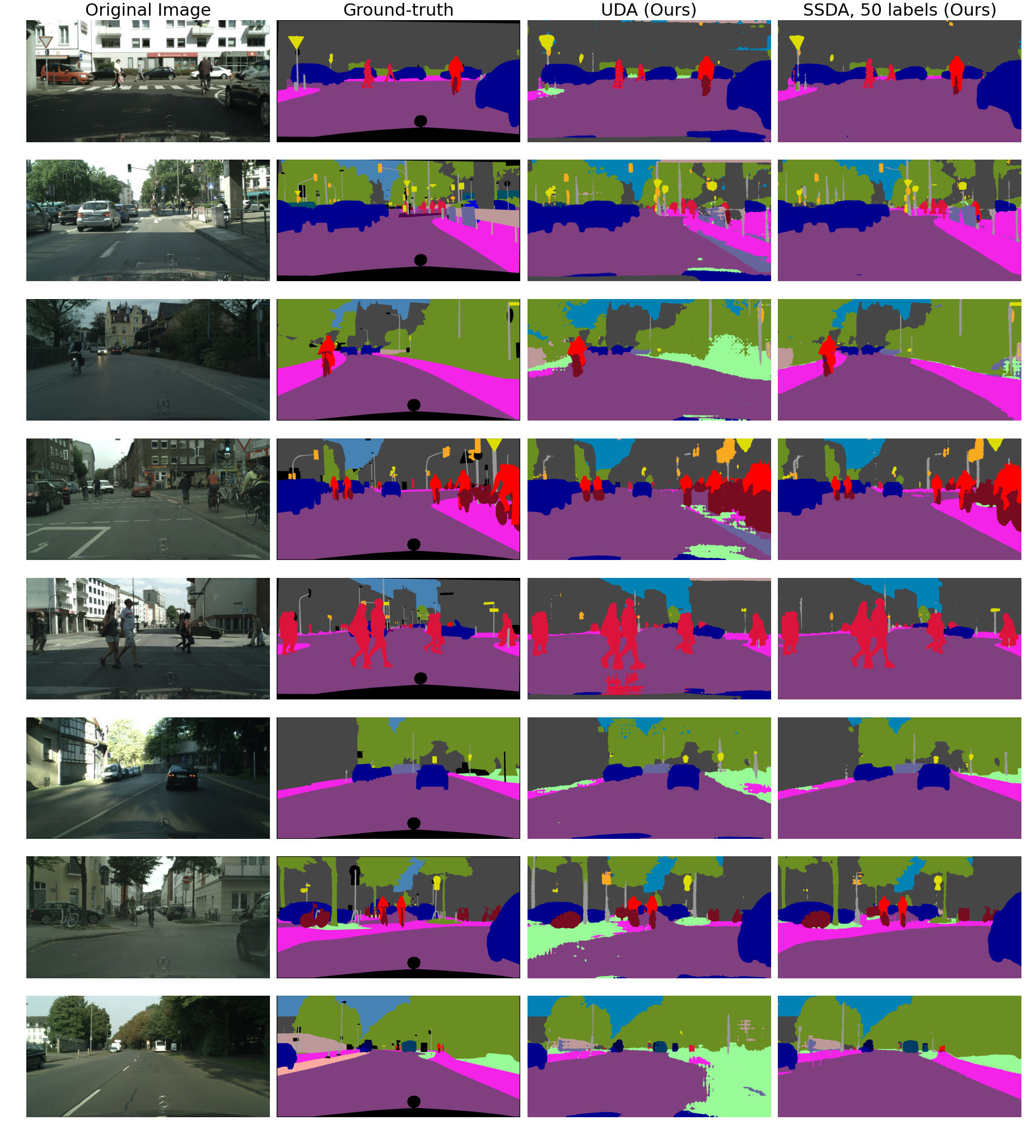}
    \caption{Qualitative results on validation images from Cityscapes. From left to right, original image used as input, ground-truth segmentation label, prediction by a model trained with our framework in a  UDA setting, and prediction by a model trained with our framework in an SSDA setting with 50 target labels. We observe a substantial qualitative improvement in the predictions of the model trained on SSDA, confirming the improvement also observed in quantitative results.}
    \label{fig:example_results}
\end{figure}

\end{document}

%% file: math_commands.tex
\usepackage{amsmath,amsfonts,bm}

\def\eqref#1{equation~\ref{#1}}

\def\1{\bm{1}}

\DeclareMathAlphabet{\mathsfit}{\encodingdefault}{\sfdefault}{m}{sl}
\SetMathAlphabet{\mathsfit}{bold}{\encodingdefault}{\sfdefault}{bx}{n}













%% file: main.bbl
\begin{thebibliography}{59}
\providecommand{\natexlab}[1]{#1}
\providecommand{\url}[1]{\texttt{#1}}
\expandafter\ifx\csname urlstyle\endcsname\relax
  \providecommand{\doi}[1]{doi: #1}\else
  \providecommand{\doi}{doi: \begingroup \urlstyle{rm}\Url}\fi

\bibitem[Alonso et~al.(2021)Alonso, Sabater, Ferstl, Montesano, and
  Murillo]{AlonsoSSL}
Inigo Alonso, Alberto Sabater, David Ferstl, Luis Montesano, and Ana~C Murillo.
\newblock Semi-supervised semantic segmentation with pixel-level contrastive
  learning from a class-wise memory bank.
\newblock In \emph{International Conference on Computer Vision (ICCV)}, pp.\
  8219--8228, 2021.

\bibitem[Araslanov \& Roth(2021)Araslanov and Roth]{araslanov2021self}
Nikita Araslanov and Stefan Roth.
\newblock Self-supervised augmentation consistency for adapting semantic
  segmentation.
\newblock In \emph{Conference on Computer Vision and Pattern Recognition
  (CVPR)}, pp.\  15384--15394, 2021.

\bibitem[Badrinarayanan et~al.(2017)Badrinarayanan, Kendall, and
  Cipolla]{badrinarayanan2017segnet}
Vijay Badrinarayanan, Alex Kendall, and Roberto Cipolla.
\newblock Segnet: A deep convolutional encoder-decoder architecture for image
  segmentation.
\newblock \emph{{IEEE} Transactions on Pattern Analysis and Machine
  Intelligence (T-PAMI)}, 39\penalty0 (12):\penalty0 2481--2495, 2017.

\bibitem[Berthelot et~al.(2021)Berthelot, Roelofs, Sohn, Carlini, and
  Kurakin]{Berthelot_adamatch}
David Berthelot, Rebecca Roelofs, Kihyuk Sohn, Nicholas Carlini, and Alex
  Kurakin.
\newblock Adamatch: A unified approach to semi-supervised learning and domain
  adaptation.
\newblock \emph{International Conference on Learning Representations (ICLR)},
  2021.

\bibitem[Chapelle \& Zien(2005)Chapelle and Zien]{cluster_assumption}
Olivier Chapelle and Alexander Zien.
\newblock Semi-supervised classification by low density separation.
\newblock In \emph{International workshop on artificial intelligence and
  statistics}, pp.\  57--64. PMLR, 2005.

\bibitem[Chen et~al.(2017{\natexlab{a}})Chen, Papandreou, Kokkinos, Murphy, and
  Yuille]{Chen2017deeplabv2}
Liang-Chieh Chen, George Papandreou, Iasonas Kokkinos, Kevin Murphy, and Alan~L
  Yuille.
\newblock Deeplab: Semantic image segmentation with deep convolutional nets,
  atrous convolution, and fully connected crfs.
\newblock \emph{{IEEE} Transactions on Pattern Analysis and Machine
  Intelligence (T-PAMI)}, 40\penalty0 (4):\penalty0 834--848,
  2017{\natexlab{a}}.

\bibitem[Chen et~al.(2017{\natexlab{b}})Chen, Papandreou, Kokkinos, Murphy, and
  Yuille]{deeplabv2}
Liang-Chieh Chen, George Papandreou, Iasonas Kokkinos, Kevin Murphy, and Alan~L
  Yuille.
\newblock Deeplab: Semantic image segmentation with deep convolutional nets,
  atrous convolution, and fully connected crfs.
\newblock \emph{{IEEE} Transactions on Pattern Analysis and Machine
  Intelligence (T-PAMI)}, 40\penalty0 (4):\penalty0 834--848,
  2017{\natexlab{b}}.

\bibitem[Chen et~al.(2018)Chen, Zhu, Papandreou, Schroff, and Adam]{deeplabv3}
Liang-Chieh Chen, Yukun Zhu, George Papandreou, Florian Schroff, and Hartwig
  Adam.
\newblock Encoder-decoder with atrous separable convolution for semantic image
  segmentation.
\newblock In \emph{European Conference on Computer Vision (ECCV)}, pp.\
  801--818, 2018.

\bibitem[Chen et~al.(2022)Chen, Wei, Jin, Chen, Zheng, Chen, and
  Jin]{chen2022dbb}
Lin Chen, Zhixiang Wei, Xin Jin, Huaian Chen, Miao Zheng, Kai Chen, and Yi~Jin.
\newblock Deliberated domain bridging for domain adaptive semantic
  segmentation.
\newblock \emph{Advances in neural information processing systems (NeurIPS)},
  2022.

\bibitem[Chen et~al.(2021{\natexlab{a}})Chen, Jia, He, Shi, and Liu]{ChenDual}
Shuaijun Chen, Xu~Jia, Jianzhong He, Yongjie Shi, and Jianzhuang Liu.
\newblock Semi-supervised domain adaptation based on dual-level domain mixing
  for semantic segmentation.
\newblock In \emph{Conference on Computer Vision and Pattern Recognition
  (CVPR)}, pp.\  11018--11027, 2021{\natexlab{a}}.

\bibitem[Chen et~al.(2021{\natexlab{b}})Chen, Yuan, Zeng, and
  Wang]{chen2021CPS}
Xiaokang Chen, Yuhui Yuan, Gang Zeng, and Jingdong Wang.
\newblock Semi-supervised semantic segmentation with cross pseudo supervision.
\newblock In \emph{Conference on Computer Vision and Pattern Recognition
  (CVPR)}, pp.\  2613--2622, 2021{\natexlab{b}}.

\bibitem[Cubuk et~al.(2020)Cubuk, Zoph, Shlens, and Le]{Cubuk2020randaugment}
Ekin~D Cubuk, Barret Zoph, Jonathon Shlens, and Quoc~V Le.
\newblock Randaugment: Practical automated data augmentation with a reduced
  search space.
\newblock In \emph{International Conference on Computer Vision (ICCV)}, pp.\
  702--703, 2020.

\bibitem[Fan et~al.(2022)Fan, Gao, Jin, and Jiang]{fan2022ucc}
Jiashuo Fan, Bin Gao, Huan Jin, and Lihui Jiang.
\newblock Ucc: Uncertainty guided cross-head co-training for semi-supervised
  semantic segmentation.
\newblock In \emph{Conference on Computer Vision and Pattern Recognition
  (CVPR)}, pp.\  9947--9956, 2022.

\bibitem[French et~al.(2019)French, Laine, Aila, Mackiewicz, and
  Finlayson]{french2019cutmix}
Geoff French, Samuli Laine, Timo Aila, Michal Mackiewicz, and Graham Finlayson.
\newblock Semi-supervised semantic segmentation needs strong, varied
  perturbations.
\newblock \emph{arXiv preprint arXiv:1906.01916}, 2019.

\bibitem[Ganin et~al.(2016)Ganin, Ustinova, Ajakan, Germain, Larochelle,
  Laviolette, Marchand, and Lempitsky]{Ganin_DANN}
Yaroslav Ganin, Evgeniya Ustinova, Hana Ajakan, Pascal Germain, Hugo
  Larochelle, Fran{\c{c}}ois Laviolette, Mario Marchand, and Victor Lempitsky.
\newblock Domain-adversarial training of neural networks.
\newblock \emph{Journal of Machine Learning Research}, 17\penalty0
  (1):\penalty0 2096--2030, 2016.

\bibitem[He et~al.(2021)He, Jia, Chen, and Liu]{he2021lab}
Jianzhong He, Xu~Jia, Shuaijun Chen, and Jianzhuang Liu.
\newblock Multi-source domain adaptation with collaborative learning for
  semantic segmentation.
\newblock In \emph{Conference on Computer Vision and Pattern Recognition
  (CVPR)}, pp.\  11008--11017, 2021.

\bibitem[Hoffman et~al.(2018)Hoffman, Tzeng, Park, Zhu, Isola, Saenko, Efros,
  and Darrell]{Hoffman_Cycada}
Judy Hoffman, Eric Tzeng, Taesung Park, Jun-Yan Zhu, Phillip Isola, Kate
  Saenko, Alexei Efros, and Trevor Darrell.
\newblock Cycada: Cycle-consistent adversarial domain adaptation.
\newblock In \emph{International Conference on Machine Learning (ICML)}, pp.\
  1989--1998. PMLR, 2018.

\bibitem[Hoyer et~al.(2021{\natexlab{a}})Hoyer, Dai, and
  Van~Gool]{Hoyer_DAFormer}
Lukas Hoyer, Dengxin Dai, and Luc Van~Gool.
\newblock Daformer: Improving network architectures and training strategies for
  domain-adaptive semantic segmentation.
\newblock 2021{\natexlab{a}}.

\bibitem[Hoyer et~al.(2021{\natexlab{b}})Hoyer, Dai, Wang, Chen, and
  Van~Gool]{Hoyer_SDE}
Lukas Hoyer, Dengxin Dai, Qin Wang, Yuhua Chen, and Luc Van~Gool.
\newblock Improving semi-supervised and domain-adaptive semantic segmentation
  with self-supervised depth estimation.
\newblock \emph{arXiv preprint arXiv:2108.12545}, 2021{\natexlab{b}}.

\bibitem[Hoyer et~al.(2022)Hoyer, Dai, and Van~Gool]{Hoyer_HRDA}
Lukas Hoyer, Dengxin Dai, and Luc Van~Gool.
\newblock Hrda: Context-aware high-resolution domain-adaptive semantic
  segmentation.
\newblock \emph{European Conference on Computer Vision (ECCV)}, 2022.

\bibitem[Ke et~al.(2020)Ke, Qiu, Li, Yan, and Lau]{ke2020guided}
Zhanghan Ke, Di~Qiu, Kaican Li, Qiong Yan, and Rynson~WH Lau.
\newblock Guided collaborative training for pixel-wise semi-supervised
  learning.
\newblock In \emph{European Conference on Computer Vision (ECCV)}, pp.\
  429--445. Springer, 2020.

\bibitem[Kim \& Kim(2020)Kim and Kim]{kim2020attract}
Taekyung Kim and Changick Kim.
\newblock Attract, perturb, and explore: Learning a feature alignment network
  for semi-supervised domain adaptation.
\newblock In \emph{European Conference on Computer Vision (ECCV)}, pp.\
  591--607. Springer, 2020.

\bibitem[Kwon \& Kwak(2022)Kwon and Kwak]{kwon2022ELN}
Donghyeon Kwon and Suha Kwak.
\newblock Semi-supervised semantic segmentation with error localization
  network.
\newblock In \emph{Conference on Computer Vision and Pattern Recognition
  (CVPR)}, pp.\  9957--9967, 2022.

\bibitem[Li et~al.(2022{\natexlab{a}})Li, Wang, Gao, and Hu]{li2022ehtdi}
Junjie Li, Zilei Wang, Yuan Gao, and Xiaoming Hu.
\newblock Exploring high-quality target domain information for unsupervised
  domain adaptive semantic segmentation.
\newblock In \emph{Proceedings of the 30th ACM International Conference on
  Multimedia}, pp.\  5237--5245, 2022{\natexlab{a}}.

\bibitem[Li et~al.(2022{\natexlab{b}})Li, Li, He, Zhang, Jia, and
  Zhang]{li2022cpsl}
Ruihuang Li, Shuai Li, Chenhang He, Yabin Zhang, Xu~Jia, and Lei Zhang.
\newblock Class-balanced pixel-level self-labeling for domain adaptive semantic
  segmentation.
\newblock In \emph{Conference on Computer Vision and Pattern Recognition
  (CVPR)}, pp.\  11593--11603, 2022{\natexlab{b}}.

\bibitem[Li et~al.(2019)Li, Yuan, and Vasconcelos]{Li2019bidirectional}
Yunsheng Li, Lu~Yuan, and Nuno Vasconcelos.
\newblock Bidirectional learning for domain adaptation of semantic
  segmentation.
\newblock In \emph{Conference on Computer Vision and Pattern Recognition
  (CVPR)}, pp.\  6936--6945, 2019.

\bibitem[Liu et~al.(2022{\natexlab{a}})Liu, Liu, Zhu, Shen, and
  Fernandez-Granda]{liu2022adaptive}
Sheng Liu, Kangning Liu, Weicheng Zhu, Yiqiu Shen, and Carlos Fernandez-Granda.
\newblock Adaptive early-learning correction for segmentation from noisy
  annotations.
\newblock In \emph{Conference on Computer Vision and Pattern Recognition
  (CVPR)}, pp.\  2606--2616, 2022{\natexlab{a}}.

\bibitem[Liu et~al.(2022{\natexlab{b}})Liu, Zhi, Johns, and
  Davison]{liu2022reco}
Shikun Liu, Shuaifeng Zhi, Edward Johns, and Andrew~J Davison.
\newblock Bootstrapping semantic segmentation with regional contrast.
\newblock In \emph{International Conference on Learning Representations
  (ICLR)}, 2022{\natexlab{b}}.

\bibitem[Liu et~al.(2021)Liu, Deng, Gao, Li, and Duan]{liu2021bapa}
Yahao Liu, Jinhong Deng, Xinchen Gao, Wen Li, and Lixin Duan.
\newblock Bapa-net: Boundary adaptation and prototype alignment for
  cross-domain semantic segmentation.
\newblock In \emph{International Conference on Computer Vision (ICCV)}, pp.\
  8801--8811, 2021.

\bibitem[Liu et~al.(2022{\natexlab{c}})Liu, Tian, Chen, Liu, Belagiannis, and
  Carneiro]{liu2022perturbed}
Yuyuan Liu, Yu~Tian, Yuanhong Chen, Fengbei Liu, Vasileios Belagiannis, and
  Gustavo Carneiro.
\newblock Perturbed and strict mean teachers for semi-supervised semantic
  segmentation.
\newblock In \emph{Conference on Computer Vision and Pattern Recognition
  (CVPR)}, pp.\  4258--4267, 2022{\natexlab{c}}.

\bibitem[Mei et~al.(2020)Mei, Zhu, Zou, and Zhang]{Mei_IAST}
Ke~Mei, Chuang Zhu, Jiaqi Zou, and Shanghang Zhang.
\newblock Instance adaptive self-training for unsupervised domain adaptation.
\newblock In \emph{European Conference on Computer Vision (ECCV)}, pp.\
  415--430. Springer, 2020.

\bibitem[Mishra et~al.(2021)Mishra, Saenko, and Saligrama]{Mishra_PAC}
Samarth Mishra, Kate Saenko, and Venkatesh Saligrama.
\newblock Surprisingly simple semi-supervised domain adaptation with
  pretraining and consistency.
\newblock \emph{In Proceedings of the British Machine Vision Conference}, 2021.

\bibitem[Olsson et~al.(2021)Olsson, Tranheden, Pinto, and
  Svensson]{olsson2021classmix}
Viktor Olsson, Wilhelm Tranheden, Juliano Pinto, and Lennart Svensson.
\newblock Classmix: Segmentation-based data augmentation for semi-supervised
  learning.
\newblock In \emph{Proceedings of the IEEE/CVF Winter Conference on
  Applications of Computer Vision}, pp.\  1369--1378, 2021.

\bibitem[Pissas et~al.(2022)Pissas, Ravasio, Da~Cruz, and
  Bergeles]{pissas2022multi}
Theodoros Pissas, Claudio~S Ravasio, Lyndon Da~Cruz, and Christos Bergeles.
\newblock Multi-scale and cross-scale contrastive learning for semantic
  segmentation.
\newblock 2022.

\bibitem[Qin et~al.(2021)Qin, Wang, Ma, Yin, Wang, and Fu]{Qin2021uoda}
Can Qin, Lichen Wang, Qianqian Ma, Yu~Yin, Huan Wang, and Yun Fu.
\newblock Contradictory structure learning for semi-supervised domain
  adaptation.
\newblock In \emph{Proceedings of the 2021 SIAM International Conference on
  Data Mining (SDM)}, pp.\  576--584. SIAM, 2021.

\bibitem[Richter et~al.(2016)Richter, Vineet, Roth, and
  Koltun]{Richter2016gta5}
Stephan~R Richter, Vibhav Vineet, Stefan Roth, and Vladlen Koltun.
\newblock Playing for data: Ground truth from computer games.
\newblock In \emph{European Conference on Computer Vision (ECCV)}, pp.\
  102--118. Springer, 2016.

\bibitem[Richter et~al.(2022)Richter, Al~Haija, and
  Koltun]{richter2022photorealism}
Stephan~R Richter, Hassan~Abu Al~Haija, and Vladlen Koltun.
\newblock Enhancing photorealism enhancement.
\newblock \emph{{IEEE} Transactions on Pattern Analysis and Machine
  Intelligence (T-PAMI)}, 2022.

\bibitem[Ronneberger et~al.(2015)Ronneberger, Fischer, and
  Brox]{ronneberger2015u}
Olaf Ronneberger, Philipp Fischer, and Thomas Brox.
\newblock U-net: Convolutional networks for biomedical image segmentation.
\newblock In \emph{International Conference on Medical image computing and
  computer-assisted intervention}, pp.\  234--241. Springer, 2015.

\bibitem[Ros et~al.(2016)Ros, Sellart, Materzynska, Vazquez, and
  Lopez]{ros2016synthia}
German Ros, Laura Sellart, Joanna Materzynska, David Vazquez, and Antonio~M
  Lopez.
\newblock The synthia dataset: A large collection of synthetic images for
  semantic segmentation of urban scenes.
\newblock In \emph{Conference on Computer Vision and Pattern Recognition
  (CVPR)}, pp.\  3234--3243, 2016.

\bibitem[Saito et~al.(2019)Saito, Kim, Sclaroff, Darrell, and
  Saenko]{Saito2019MME}
Kuniaki Saito, Donghyun Kim, Stan Sclaroff, Trevor Darrell, and Kate Saenko.
\newblock Semi-supervised domain adaptation via minimax entropy.
\newblock In \emph{International Conference on Computer Vision (ICCV)}, pp.\
  8050--8058, 2019.

\bibitem[Saito et~al.(2021)Saito, Kim, Teterwak, Sclaroff, Darrell, and
  Saenko]{saito2021tune}
Kuniaki Saito, Donghyun Kim, Piotr Teterwak, Stan Sclaroff, Trevor Darrell, and
  Kate Saenko.
\newblock Tune it the right way: Unsupervised validation of domain adaptation
  via soft neighborhood density.
\newblock In \emph{International Conference on Computer Vision (ICCV)}, pp.\
  9184--9193, 2021.

\bibitem[Sohn et~al.(2020)Sohn, Berthelot, Carlini, Zhang, Zhang, Raffel,
  Cubuk, Kurakin, and Li]{sohn2020fixmatch}
Kihyuk Sohn, David Berthelot, Nicholas Carlini, Zizhao Zhang, Han Zhang,
  Colin~A Raffel, Ekin~Dogus Cubuk, Alexey Kurakin, and Chun-Liang Li.
\newblock Fixmatch: Simplifying semi-supervised learning with consistency and
  confidence.
\newblock \emph{Advances in neural information processing systems (NeurIPS)},
  33:\penalty0 596--608, 2020.

\bibitem[Tarvainen \& Valpola(2017)Tarvainen and Valpola]{MeanTeacher}
Antti Tarvainen and Harri Valpola.
\newblock Mean teachers are better role models: Weight-averaged consistency
  targets improve semi-supervised deep learning results.
\newblock \emph{Advances in neural information processing systems (NeurIPS)},
  30, 2017.

\bibitem[Teh et~al.(2022)Teh, DeVries, Duke, Jiang, Aarabi, and
  Taylor]{teh2022gist}
Eu~Wern Teh, Terrance DeVries, Brendan Duke, Ruowei Jiang, Parham Aarabi, and
  Graham~W Taylor.
\newblock The gist and rist of iterative self-training for semi-supervised
  segmentation.
\newblock In \emph{2022 19th Conference on Robots and Vision (CRV)}, pp.\
  58--66. IEEE, 2022.

\bibitem[Tranheden et~al.(2021)Tranheden, Olsson, Pinto, and
  Svensson]{Tranheden_dacs}
Wilhelm Tranheden, Viktor Olsson, Juliano Pinto, and Lennart Svensson.
\newblock Dacs: Domain adaptation via cross-domain mixed sampling.
\newblock In \emph{Proceedings of the IEEE/CVF Winter Conference on
  Applications of Computer Vision}, pp.\  1379--1389, 2021.

\bibitem[Tu et~al.(2022)Tu, Huang, Ji, Zheng, and Shao]{Tu2022guidedmix}
Peng Tu, Yawen Huang, Rongrong Ji, Feng Zheng, and Ling Shao.
\newblock Guidedmix-net: Learning to improve pseudo masks using labeled images
  as reference.
\newblock \emph{AAAI Conference on Artificial Intelligence}, 2022.

\bibitem[Vu et~al.(2019)Vu, Jain, Bucher, Cord, and P{\'e}rez]{Vu_AdvEnt}
Tuan-Hung Vu, Himalaya Jain, Maxime Bucher, Matthieu Cord, and Patrick
  P{\'e}rez.
\newblock Advent: Adversarial entropy minimization for domain adaptation in
  semantic segmentation.
\newblock In \emph{Conference on Computer Vision and Pattern Recognition
  (CVPR)}, pp.\  2517--2526, 2019.

\bibitem[Wang et~al.(2020{\natexlab{a}})Wang, Shen, Zhang, Duan, and
  Mei]{wang2020classes}
Haoran Wang, Tong Shen, Wei Zhang, Ling-Yu Duan, and Tao Mei.
\newblock Classes matter: A fine-grained adversarial approach to cross-domain
  semantic segmentation.
\newblock In \emph{European Conference on Computer Vision (ECCV)}, pp.\
  642--659. Springer, 2020{\natexlab{a}}.

\bibitem[Wang et~al.(2021)Wang, Zhou, Yu, Dai, Konukoglu, and
  Van~Gool]{WangPCL}
Wenguan Wang, Tianfei Zhou, Fisher Yu, Jifeng Dai, Ender Konukoglu, and Luc
  Van~Gool.
\newblock Exploring cross-image pixel contrast for semantic segmentation.
\newblock In \emph{International Conference on Computer Vision (ICCV)}, pp.\
  7303--7313, 2021.

\bibitem[Wang et~al.(2020{\natexlab{b}})Wang, Wei, Feris, Xiong, Hwu, Huang,
  and Shi]{Wang_ASS}
Zhonghao Wang, Yunchao Wei, Rogerio Feris, Jinjun Xiong, Wen-Mei Hwu, Thomas~S
  Huang, and Honghui Shi.
\newblock Alleviating semantic-level shift: A semi-supervised domain adaptation
  method for semantic segmentation.
\newblock In \emph{Conference on Computer Vision and Pattern Recognition
  (CVPR)}, pp.\  936--937, 2020{\natexlab{b}}.

\bibitem[Xie et~al.(2021)Xie, Wang, Yu, Anandkumar, Alvarez, and
  Luo]{xie2021segformer}
Enze Xie, Wenhai Wang, Zhiding Yu, Anima Anandkumar, Jose~M Alvarez, and Ping
  Luo.
\newblock Segformer: Simple and efficient design for semantic segmentation with
  transformers.
\newblock \emph{Advances in neural information processing systems (NeurIPS)},
  34:\penalty0 12077--12090, 2021.

\bibitem[Xie et~al.(2020)Xie, Luong, Hovy, and Le]{xie2020noisy_student}
Qizhe Xie, Minh-Thang Luong, Eduard Hovy, and Quoc~V Le.
\newblock Self-training with noisy student improves imagenet classification.
\newblock In \emph{Conference on Computer Vision and Pattern Recognition
  (CVPR)}, pp.\  10687--10698, 2020.

\bibitem[Yang \& Soatto(2020)Yang and Soatto]{Yang_FDA}
Yanchao Yang and Stefano Soatto.
\newblock Fda: Fourier domain adaptation for semantic segmentation.
\newblock In \emph{Conference on Computer Vision and Pattern Recognition
  (CVPR)}, pp.\  4085--4095, 2020.

\bibitem[Yu et~al.(2020)Yu, Chen, Wang, Xian, Chen, Liu, Madhavan, and
  Darrell]{yu2020bdd100k}
Fisher Yu, Haofeng Chen, Xin Wang, Wenqi Xian, Yingying Chen, Fangchen Liu,
  Vashisht Madhavan, and Trevor Darrell.
\newblock Bdd100k: A diverse driving dataset for heterogeneous multitask
  learning.
\newblock In \emph{Conference on Computer Vision and Pattern Recognition
  (CVPR)}, pp.\  2636--2645, 2020.

\bibitem[Yun et~al.(2019)Yun, Han, Oh, Chun, Choe, and Yoo]{yun2019cutmix}
Sangdoo Yun, Dongyoon Han, Seong~Joon Oh, Sanghyuk Chun, Junsuk Choe, and
  Youngjoon Yoo.
\newblock Cutmix: Regularization strategy to train strong classifiers with
  localizable features.
\newblock In \emph{International Conference on Computer Vision (ICCV)}, pp.\
  6023--6032, 2019.

\bibitem[Zhang et~al.(2021)Zhang, Zhang, Zhang, Chen, Wang, and
  Wen]{Zhang_proda}
Pan Zhang, Bo~Zhang, Ting Zhang, Dong Chen, Yong Wang, and Fang Wen.
\newblock Prototypical pseudo label denoising and target structure learning for
  domain adaptive semantic segmentation.
\newblock In \emph{Conference on Computer Vision and Pattern Recognition
  (CVPR)}, pp.\  12414--12424, 2021.

\bibitem[Zoph et~al.(2020)Zoph, Ghiasi, Lin, Cui, Liu, Cubuk, and
  Le]{zoph2020rethinking}
Barret Zoph, Golnaz Ghiasi, Tsung-Yi Lin, Yin Cui, Hanxiao Liu, Ekin~Dogus
  Cubuk, and Quoc Le.
\newblock Rethinking pre-training and self-training.
\newblock \emph{Advances in neural information processing systems (NeurIPS)},
  33:\penalty0 3833--3845, 2020.

\bibitem[Zou et~al.(2018)Zou, Yu, Kumar, and Wang]{zou2018cbst}
Yang Zou, Zhiding Yu, BVK Kumar, and Jinsong Wang.
\newblock Unsupervised domain adaptation for semantic segmentation via
  class-balanced self-training.
\newblock In \emph{European Conference on Computer Vision (ECCV)}, pp.\
  289--305, 2018.

\bibitem[Zou et~al.(2020)Zou, Zhang, Zhang, Li, Bian, Huang, and
  Pfister]{zou2020pseudoseg}
Yuliang Zou, Zizhao Zhang, Han Zhang, Chun-Liang Li, Xiao Bian, Jia-Bin Huang,
  and Tomas Pfister.
\newblock Pseudoseg: Designing pseudo labels for semantic segmentation.
\newblock \emph{arXiv preprint arXiv:2010.09713}, 2020.

\end{thebibliography}
